\documentclass[journal]{IEEEtran}
\usepackage{bbding}
\usepackage{color}
\makeatletter
\newcommand\figcaption{\def\@captype{figure}\caption}
\newcommand\tabcaption{\def\@captype{table}\caption}
\makeatother
\usepackage{multirow}
\AtBeginDocument{%
  \providecommand\BibTeX{{%
    \normalfont B\kern-0.5em{\scshape i\kern-0.25em b}\kern-0.8em\TeX}}}

\ifCLASSINFOpdf
   \usepackage[pdftex]{graphicx}
\else
\fi
\usepackage{array}
\usepackage{amsmath}
\usepackage{subfigure}
\begin{document}
%
\title{Shadow-Aware Dynamic Convolution for Shadow Removal}
%
%
%

\author{Yimin~Xu,
        Mingbao~Lin,
        Hong~Yang,
        Fei Chao,
        Rongrong Ji,~\IEEEmembership{Senior Member, IEEE}
\thanks{This work was supported by the National Science Fund for Distinguished Young Scholars (No. 62025603), the National Natural Science Foundation of China (No. U21B2037, No. 62176222, No. 62176223, No. 62176226, No. 62072386, No. 62072387, No. 62072389, and No. 62002305), Guangdong Basic and Applied Basic Research Foundation (No. 2019B1515120049), and the Natural Science Foundation of Fujian Province of China (No. 2021J01002).}
\thanks{Y. Xu is with Media Analytics and Computing Laboratory, Department of Artificial Intelligence, School of Informatics, Xiamen University, Xiamen 361005, China.}
\thanks{M. Lin is with the Tencent Youtu Lab, Shanghai 200233, China.}
\thanks{H. Yang, F. Chao and R. Ji (Corresponding Author) are with the Media Analytics and Computing Laboratory, Department of Artificial Intelligence, School of Informatics, Xiamen University, Xiamen 361005, China (e-mail: rrji@xmu.edu.cn).}%
}
\maketitle

\begin{abstract}
With a wide range of shadows in many collected images, shadow removal has aroused increasing attention since uncontaminated images are of vital importance for many image processing tasks.
Current methods consider the same convolution operations for both shadow and non-shadow regions while ignoring the large gap between the color mappings for the shadow region and the non-shadow region, leading to poor quality of reconstructed images and a heavy computation burden.
To solve this problem, this paper introduces a novel plug-and-play Shadow-Aware Dynamic Convolution (SADC) module to decouple the interdependence between the shadow region and the non-shadow region. Inspired by the fact that the color mapping of the non-shadow region is easier to learn, our SADC processes the non-shadow region with a lightweight convolution module in a computationally cheap manner and recovers the shadow region with a more complicated convolution module to ensure the quality of image reconstruction.
Given that the non-shadow region often contains more background color information, we further develop a novel intra-convolution distillation loss to strengthen the information flow from the non-shadow region to the shadow region. 
Extensive experiments on the ISTD and SRD datasets show our method achieves better performance in shadow removal over many state-of-the-arts.
Code has been made available at https://github.com/xuyimin0926/SADC.
\end{abstract}


%
\IEEEpeerreviewmaketitle

\section{Introduction}\label{sec:introduction}
%
%
%
%
\IEEEPARstart{C}onvolutional neural networks (CNNs) has become dominant in image processing field, such as image restoration~\cite{liang2021swinir,zamir2022restormer}, image reconstruction~\cite{jiang2021focal,gu2019scene}, and image enhancement~\cite{hong1998fingerprint,lv2021attention}. To achieve better results in these tasks, a solid training set is much-needed with noise-free input samples and accurate labels. However, as shown in the recent study~\cite{akhtar2018threat}, a slight perturbation in training samples misleads the CNNs towards incorrect predictions. 
Image shadow, one of the most common perturbations in daily lives, frequently occurs when the light source is partially or fully blocked. The appearance of shadows may blur or even invisiblize vital information in an image, which hinders the further usage of collected images in many downstream image processing tasks.

\begin{figure}[!t]
    \centering
    \includegraphics[width=\linewidth]{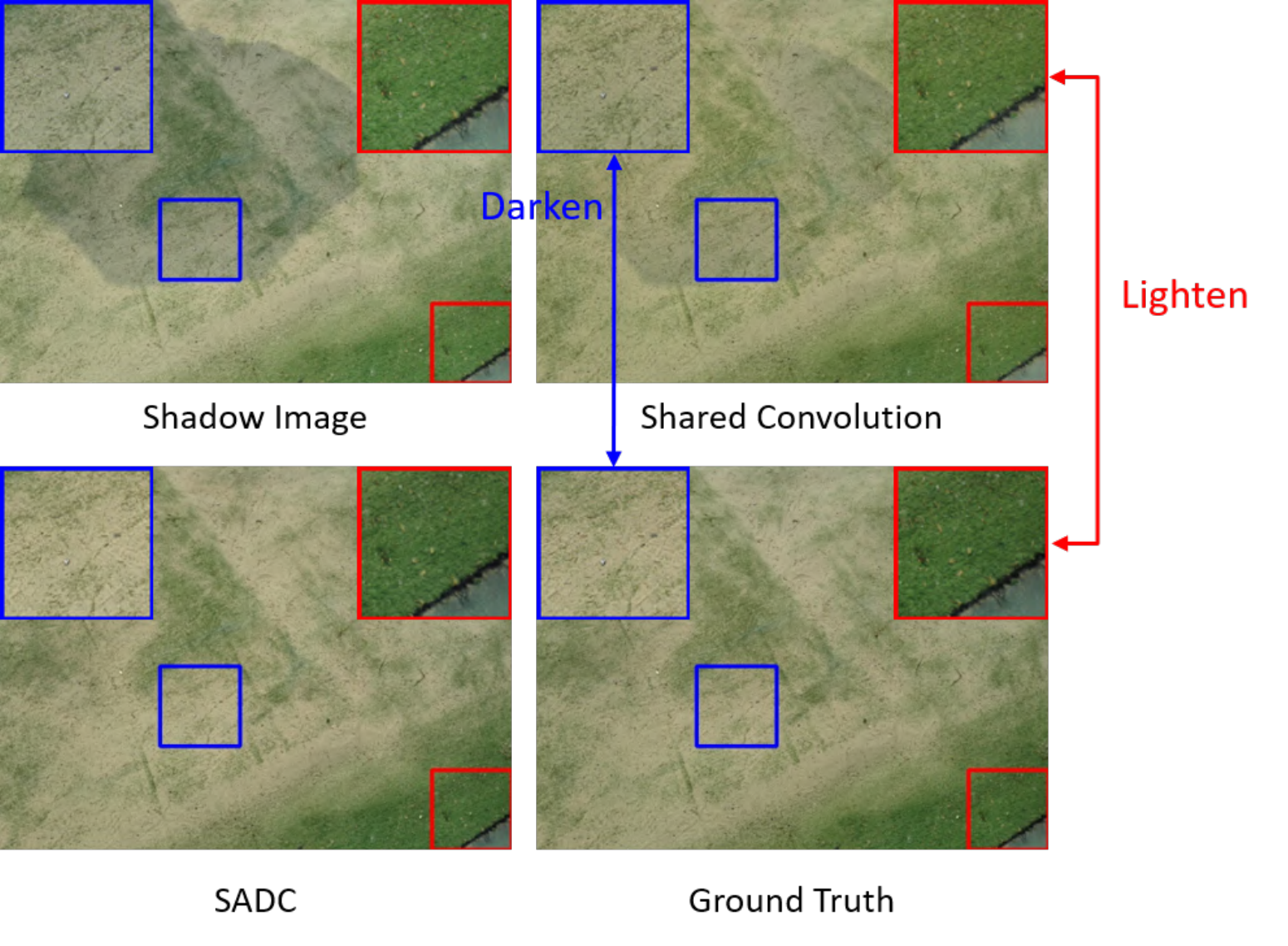}
    \caption{Contradiction in color mapping between shadow region (blue rectangle) and non-shadow region (red rectangle). The results with the shared convolution and the proposed SADC module are tested on the same backbone as shown in Fig.\,\ref{fig:fullnet}.}
    \label{fig:introduction}
\end{figure}

To obtain shadow-free images, two mainstreams in shadow removal, including physical-based methods and learning-based methods, have been developed over the past few decades.
In the early days, researchers tried to adopt a physical model to analyze each pixel's intensity~\cite{guo2012paired,gong2016interactive}. However, these methods fail to achieve appreciable shadow removal results because few information is available to find an appropriate correlation between polluted areas and their ground-truth counterparts. As an alternative, learning-based methods~\cite{vicente2017leave,nair2019shadow} accord with the data-driven manner and train a shadow removal model on the premise of abundant training samples. 
In particular, learning-based methods are becoming prosperous along with the widespread success of CNNs. Consequently, more and more researchers try to recover shadow images in a deep-learning fashion. Thanks to the higher capacity and stronger adaptability to various datasets of CNN models, plentiful works achieve substantially record-breaking performance in removing image shadows~\cite{cun2020towards,fu2021auto,le2019shadow,le2020shadow,hu2019direction, chen2021canet}.
In essence, they model a color mapping from shadow images to uncontaminated ones. 

Despite the progress, existing researches simply regard a whole shadow-contaminated image as an integral.
They deploy an identical convolution to process the input feature maps by sliding over the space dimension regardless of the varying space information between the shadow region and the non-shadow region. Consequently, two crucial issues arise in existing studies:
(1) Poor quality of reconstructed images. We realize that the optimal learned weights for the shadow region fail to perform well in the non-shadow region and vice versa. Shadow and non-shadow regions possess different color mappings from the shadow-contaminated image to its ground truth. Hence, a compromise has to be made, 
%
%
as shown in Fig.\,\ref{fig:introduction}, resulting in the over-whiten for the non-shadow region and over-darken for the shadow region. Thus, it is inappropriate to use the same convolutional operations for every region of the contaminated image.
(2) A waste of computational resources. The complexity of learning a color mapping varies a lot \emph{w.r.t}. different regions. Intuitively, modelling the color mapping in the non-shadow region is much easier to complete. Thus, it is natural to design the color mapping in a low overhead manner. However, in order to remove shadows, most existing methods construct a set of sequentially-stacked convolutional operations to process the shadow region. However, these operations are too expensive for the non-shadow region when the same convolutions are applied to the whole image.

Motivated by the above analysis, in this paper, we propose a plug-and-play Shadow-Aware Dynamic Convolution (SADC) module to decouple convolutions in the shadow region from that in the non-shadow region. 
The proposed SADC module not only separates convolutional operations between the shadow region and the non-shadow region but also re-assigns the computation cost based on the difficulty of learning a proper color mapping function. 
Considering that the color mapping for the non-shadow region is relatively straightforward due to its minor color shifting, our SADC employs a lightweight convolution module to save the computation cost while the shadow region is recovered by a more sophisticated convolution module which guarantees the high quality of reconstructed images.
Consequently, our SADC reduces the overall computation cost for the sake of deployment on on-chip devices with very limited computation resources, meanwhile, it increases the performance in both the shadow region and non-shadow region. 
As a plug-and-play method, our SADC can be easily combined with many widely-used CNNs to boost their performance in shadow removal.

As we observe in Fig.\,\ref{fig:introduction}, the non-shadow region often contains richer background information, providing a useful recovery hint to its surrounding shadow region.
Therefore, we further propose a novel intra-convolution distillation loss to convert the information flows from the non-shadow region to the shadow region.
Although processed by convolutions with different costs, our intra-convolution distillation encourages the output in the shadow region to share a similar color distribution with that in the non-shadow region, especially in the shadow boundary. Through our intra-convolution distillation, the shadow region learns the post-processed color information from the non-shadow region for restoring images.

In summary, our main contributions are three-fold:
\begin{itemize}
    \item We propose a plug-and-play shadow-aware dynamic convolution module to individually handle shadow and non-shadow regions in a computationally economical manner.
    \item We design a specific intra-convolution distillation loss to enhance the shadow removal performance based on the non-shadow convolution results.
    \item Extensive experimental results on ISTD~\cite{wang2018stacked} and SRD~\cite{qu2017deshadownet} datasets show that our SADC achieves better performance in shadow removal over many existing state-of-the-arts.
\end{itemize}

\section{Related Work}
In this section, we briefly review some studies mostly related to our work, including shadow removal~\cite{tiwari2016survey} and dynamic network~\cite{han2021dynamic}. 

\subsection{Shadow Removal}
There are two mainstream methods to conduct the single image shadow removal. Traditional approaches remove image shadow by introducing intrinsic local gradient~\cite{gryka2015learning}, illumination~\cite{xiao2013fast, zhang2018improving}, spatial correlation~\cite{guo2012paired}, \emph{etc}., as prior knowledge, and pixel values inside the shadow region are inferred by a physical model.
Recent years have witnessed the domination of CNN-based approaches due to their better performance, higher capacity, and stronger adaptability to various datasets. 
The Deshadow-Net~\cite{qu2017deshadownet} for the first time introduced an automatic multi-context deep network for shadow removal in an end-to-end learning manner. 
Cun~\emph{et al.} proposed a dual hierarchical aggregation network~\cite{cun2020towards} to detect and recover the shadow region.
Fu~\emph{et al.}~\cite{fu2021auto} regarded the shadow removal as an auto-exposure fusion problem where several over-exposure images are generated first and then fused according to auto-generated fusion weights.
CANet~\cite{chen2021canet} absorbed the contextual matching information by transferring the non-shadow features to similar shadow patches via a contextual feature transfer mechanism. 

Although these methods enhance the performance of shadow removal more or less, they ignore the contradiction of color mappings between the shadow region and the non-shadow region as detailed in Sec.\,\ref{sec:introduction}. Furthermore, the contextual mapping of CANet needs to train an over-parameterized auxiliary network to discover similar image patches. In contrast, our proposed shadow-aware dynamic convolution can enhance the overall performance by decoupling the convolutions inside the shadow region from those outside the shadow region. In addition, we propose a novel intra-convolution distillation loss without introducing a complex matching module and obtain better shadow removal results.

\begin{figure*}[!t]
\centering
\subfigure[Overall pipeline of training and testing phase of SADC]{
\begin{minipage}[t]{0.55\textwidth}
\centering
\includegraphics[width=\linewidth]{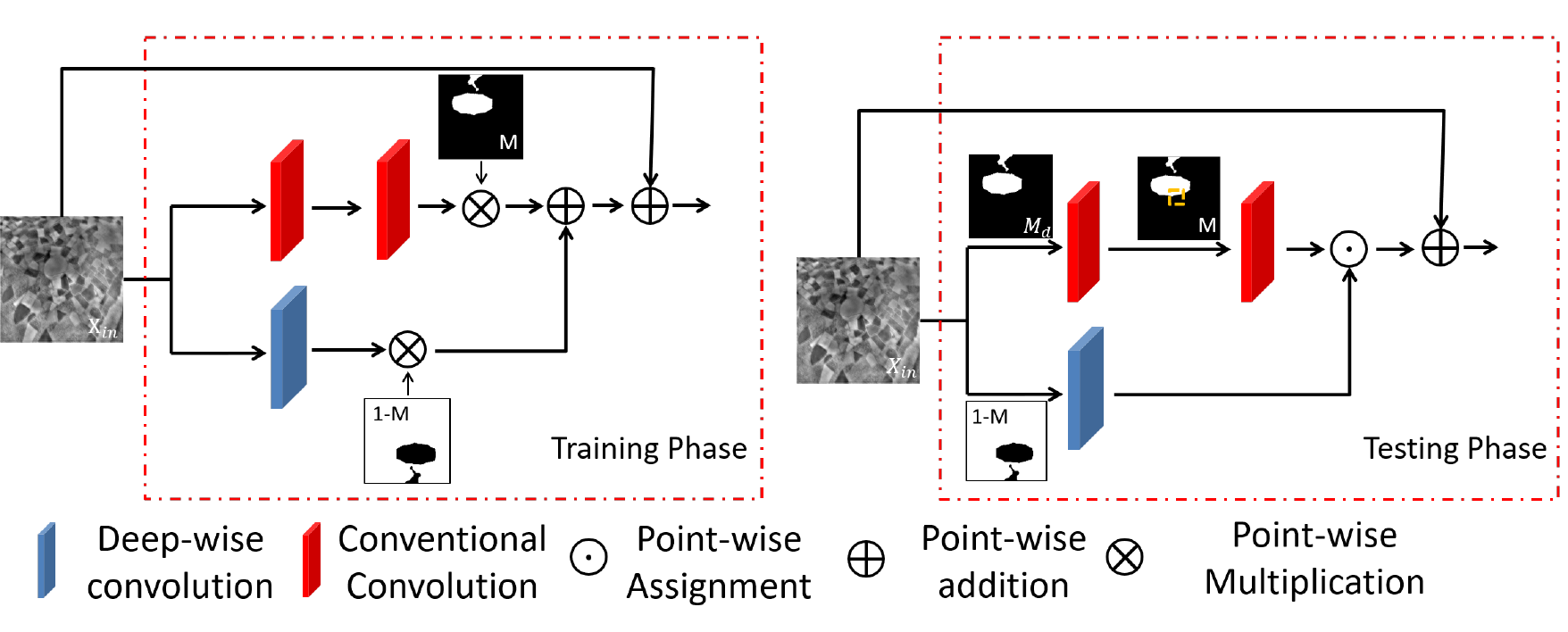}
\label{fig:SADC_two_phase}
\end{minipage}}
\subfigure[Our backbone network constructed from DHAN~\cite{cun2020towards}]{
\begin{minipage}[t]{0.42\textwidth}
\centering
\includegraphics[width=\linewidth]{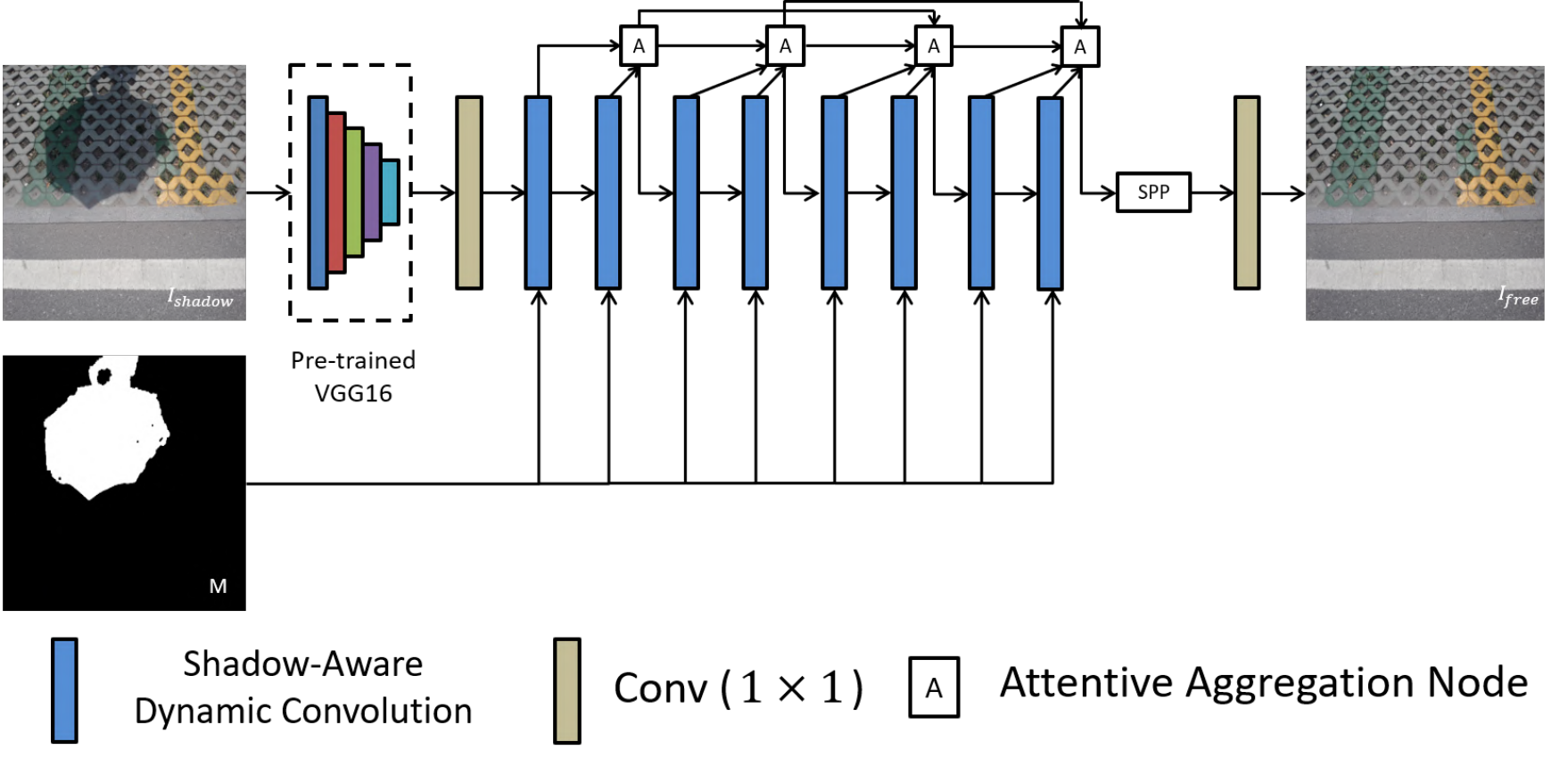}
\label{fig:fullnet}
\end{minipage}}
\caption{(a) is the overall pipeline of the training (left) and testing (right) phase of the SADC. In training phase, the whole image is sent to two branches simultaneously, while the input image is split into the shadow and non-shadow region based on the shadow mask and sent corresponding to different branches. (b) Our network structure is borrowed from DHAN~\cite{cun2020towards}. Details of the SPP and Attentive Aggregation Node can be referred to ~\cite{cun2020towards}.}
\end{figure*}

\subsection{Dynamic network}
Despite the success of CNNs, most of them adopt a static inference, where the computational graph and network parameters are fixed for all input images, which limits the representation power and thus harms their performance since different images often contain varying contents. 
In contrast, by assigning different computational weights~\cite{chen2020dynamic, li2020revisiting} or various execution paths~\cite{gao2018dynamic,li2021dynamic} to the inputs, dynamic neural networks enjoy many advantages over the conventional static CNNs in terms of inference efficiency, representation power, adaptiveness, \emph{etc}.
Huang~\emph{et al.}~\cite{huang2018multi} devised a multi-scale dense network to process samples with different resolutions and sub-networks. 
Verelst~\emph{et al.}~\cite{verelst2020dynamic} proposed a dynamic convolutions to skip the useless convolution in spatial domain. By only conducting convolution on important pixels, the network can reach classification accuracy on par with the competitors on all images.
Xie~\emph{et al.}~\cite{xie2020spatially} accomplished object detection via a spatial adaptive feature sampling and interpolation. Instead of only computing the convolutions of certain values, they estimated the convolution results of other pixels via interpolation.

Inspired by the findings in dynamic neural networks, we propose a dedicate shadow-aware dynamic convolution module specifically for the shadow removal task. Since the input samples have different shadows and the shadow area is endowed with different convolutions, our method realizes an instance-wise and spatial-wise dynamic network simultaneously. To our best knowledge, this is the first work to decouple shadow region and non-shadow region with different convolution complexities.

\section{Methods}
Our approach to shadow removal is motivated by the fact that the color mapping in the shadow region are more complex in comparison with the non-shadow region.
Inspired by this fact, 
we propose shadow-aware dynamic convolution to learn different mapping functions for shadow and non-shadow regions.
Sec.\,\ref{preliminary} gives some preliminaries.
We introduce the realization of our shadow-aware convolution in Sec.\,\ref{shadow-aware} and show its convenience for deployment on current networks. In Sec.\,\ref{intra-convolution}, we present our intra-convolution distillation loss and Sec.\,\ref{overall} details our overall learning objective.

\subsection{Preliminary}\label{preliminary}
Suppose the input feature maps for a conventional convolution layer in the shadow removal task is $X_{in} \in \mathcal{R}^{C \times H \times W}$, where $C$ is the number of input channels, $H$ and $W$ denote the height and width of input feature maps respectively. For the conventional convolution, the output $X_{out} \in \mathcal{R}^{C' \times H' \times W'}$ is derived by:
\begin{align}
    X_{out} = W \otimes X_{in},
\end{align}
where $W \in \mathcal{R}^{C' \times C \times K \times K}$ is the convolution weight, and $\otimes$ denotes the convolution operation. 
Usually, a bias term is appended after the convolution. For brevity, we omit it in this paper.
Besides, in shadow removal task, a ground-truth shadow mask $M \in \{0, 1\}^{H \times W}$ is usually accomplished with its entry $(M)_{ij} = 1$ indicating a shadow pixel in the input $X_{in}$ and 0 indicating a non-shadow pixel.

Over the past years, various shadow removal oriented network structures have been proposed~\cite{cun2020towards,qu2017deshadownet,hu2019direction}. Most of them adopt a set of sequentially-stacked static convolutions to process the input feature maps regardless of the varying space information between shadow region and non-shadow region, leading to poor quality of reconstructed image, as well as computation waste, as stated in Sec.\,\ref{sec:introduction}. 
This motivates us to design a shadow-aware dynamic convolution module as depicted in Fig.\,\ref{fig:SADC_two_phase}, which decouples the convolutions in the shadow region from the non-shadow region.
As a plug-and-play method, our SADC leads to orthogonality to existing studies concentrated on network structures. 
For an illustrative example, we directly borrow the DHAN~\cite{cun2020towards} as our network structure in this paper, details of which are given in Fig.\,\ref{fig:fullnet}.
Note that, in DHAN, the number of input channels is equal to the number of output channels, \emph{i.e.}, $C = C'$. Thus, we substitute $C'$ with $C$ in the following contents.

\subsection{Shadow-Aware Dynamic Convolution}\label{shadow-aware}
Fig.\,\ref{fig:SADC_two_phase} gives a pictorial illustration of our SADC. Its key innovation lies in the two-branch design including a lightweight convolution operator and a conventional convolution operator.
Technically, the lightweight convolution intends to model the color mapping in the non-shadow region, meanwhile, the conventional convolution aims to learn color mapping in the shadow region.
Then, the outputs of these two branches are merged and added to the input features via a shortcut to strengthen the contextual similarity and reduce the learning difficulty. Thus, the overall workflow of our SADC can be represented as:
\begin{align}
    X_{out} = X_{out}^{m} + X_{in},
\end{align}
where $X_{out}^{m}$ denotes the merging result of the two branches.
Different from the existing learning paradigm in shadow removal, our computing of $X_{out}^m$ adapts its input to pursue a better image reconstruction in the training phase and to pursue a cheaper inference computation in the testing phase.
We detail it in the following.

\textbf{Training Phase}. 
In the training phase, the whole input feature maps are fed to both network branches to conduct convolutions of different complexities. The output of the first branch with a lightweight convolution is obtained as:
\begin{align}
    X_{out}^{ns} &= f(W_l \otimes X_{in}),
\end{align}
where $f(\cdot)$ represents a nonlinear activation function and we use LeakyReLU in our implementation. The $W_l \in \mathcal{R}^{C \times 1 \times K \times K}$ represents a lightweight deep-wise convolution. 
In contrast, the shadow region often suffers from severe color distortion and thus requires a fine-grained convolution module. Therefore, we use two consecutively conventional convolutions to recover more accurate features inside the shadow region as:
\begin{align}\label{merging_out}
    X_{out}^{s} &= f\big(W_2 \otimes f(W_1 \otimes X_{in})\big),
\end{align}
where $W_1 \in \mathcal{R}^{C \times C \times K \times K}$ and $W_2 \in \mathcal{R}^{C \times C \times K \times K}$. Finally, the $X_{out}^{m}$ is obtained as:
\begin{equation}
    (X_{out}^{m})_{:, i, j} = 
\begin{cases}
 (X_{out}^{s})_{:, i, j} & (M)_{i, j} = 1, \\
 (X_{out}^{ns})_{:, i, j} & (M)_{i, j} = 0.
\end{cases}
\end{equation}

We consider the same input feature maps for both branches for two reasons:
First, the inputs may only contain shadow pixels or non-shadow pixels due to the data augmentation, such as cropping in the training samples, which leads to the training imbalance between the two branches. In contrast, with the same inputs, we observe a better training convergence in the experiment.
Second, both the lightweight convolution and conventional convolutions can slide over the border pixels, enabling an information switch between the two branches to learn better convolution weights. It also conveniences the usage of our intra-convolution distillation described in \,\ref{intra-convolution}.
With these merits, an uncontaminated image of better quality can be obtained.

\textbf{Testing Phase}. 
The well-learned convolutional weights in the training stage ensure the quality of image reconstruction.
To reach a computational reduction of the proposed method in the testing phase, the output pixel values are generated either from the shadow branch or from the non-shadow branch based on the input shadow mask $M$. The shadow mask in the testing phase can be viewed as prior knowledge, which can be extracted by a shadow detection method~\cite{zhu2021mitigating}. Technically, our SADC separates the input feature maps $X_{in}$ into two parts including a shadow region $X_{in}^s$ and a non-shadow region $X_{in}^{ns}$. The non-shadow region is obtained as:
\begin{align}
    X_{in}^{ns} = {X_{in}}_{|M = 0}.
\end{align}

Then, with the non-shadow region $X_{in}^{ns}$ as its input, the output of the first branch in the testing stage is:
\begin{equation}\label{non-shadow}
    X_{out}^{ns} = f(W_l \otimes X_{in}^{ns}).
\end{equation}

As opposed to the whole image in the training stage, the lightweight convolution is only conducted upon the non-shadow region.

As for the shadow region, one naive solution is $X_{in}^s = {X_{in}}_{|M=1}$. However, the pixel values in the boundary transition area between the shadow region and the non-shadow region, as shown as the orange square in Fig\,\ref{fig:SADC_two_phase}, come from the output of the first convolution. If only applying the ground-truth shadow mask $M$ to the first convolution in the shadow branch, the orange region loses part of the receptive fields (\emph{e.g.}, black area in the orange square).
Instead, to better align with the training process, we first apply a morphological dilation~\cite{haralick1987image} upon the shadow mask to ensure the first convolution provides enough receptive fields for the following convolution. Then, the input of the shadow region branch is derived as:
\begin{align}
    &\bar{M}_{d} = Dilation(M; \bar{G}), \\&
    X_{in}^{s} = {X_{in}}_{|\bar{M}_{d}=1}\label{shadow},
\end{align}
where $Dilation(\cdot \, ; \bar{G})$ denotes morphology dilation with a kernel $\bar{G}$ of $k' \times k'$ size in our\ implementation, where $k'$ depends on the parameters of the original convolution.
Consequently, the output of the second branch in the testing stage is:
\begin{align}
    X_{out}^{s} &= f\big(W_2 \otimes (f(W_1 \otimes X_{in}^s)_{|M=1})\big).
\end{align}

Combining Eq.\,(\ref{non-shadow}) and Eq.\,(\ref{shadow}), the final merging result $X_{out}^m$ can be obtained by Eq.\,(\ref{merging_out}).
Different from applying the convolution to the whole feature map and conducting summation according to the shadow mask, each convolution in the SADC module is only applied to the corresponding spatial pixels, leading to computational reduction while obtaining the same results.
The supplementary material provides details of our code implementation for the testing phase.

%




\textbf{Complexity}.
It is worth noting that our SADC module can re-allocate the computation resources instead of simply splitting the original convolution into two identical convolutional kernels. 
Denoting $\rho$ as the area proportion of the non-shadow region in the whole input image, the theoretical reduction rate of floating-point operation $\gamma$ of our SADC compared to the original filter are:
\begin{align*}
    \gamma &= 1 - \frac{FLOPS_{SADC}}{FLOPS_{CONV}} \\ &\approx 1 - \frac{\rho \cdot C \cdot H \cdot W \cdot K \cdot K + 2(1-\rho) \cdot C^2 \cdot H \cdot W \cdot K^2}{C^2 \cdot H \cdot W \cdot K^2} \\ &= 1 - \frac{\rho + 2(1-\rho)\cdot C}{C} \\ &\approx 2\rho - 1.
\end{align*}

Since the dilated mask often introduces quite small extra pixels compared with a common mask, we simply approximate its computation with the common mask as depicted in the first approximation. The last approximation is based on the fact that $\rho << C$ ($0 \le \rho \le 1$).
Since the non-shadow region usually takes up most part of the whole image, indicating a large value of $\rho$, our SADC achieves great computation reduction.
Sec.\,\ref{ablation} reports our testing on practical speedups of each SADC module compared with normal convolutions.

In addition to efficiency and better performance (see Sec.\,\ref{sec:experiment}), our SADC also merits its easy deployment since it can be integrated into modern CNNs-based shadow removal models. Fig.\,\ref{fig:fullnet} shows the network structure for our shadow removal, which replaces the convolution modules in the main block of DHAN~\cite{cun2020towards} with our SADC modules. Note that, we remove the shadow detection branch in DHAN to save more computation costs since our SADC module takes the shadow mask as its input which already provides the hint for the shadow area.

\subsection{Intra-convolution Distillation Loss}\label{intra-convolution}
The non-shadow region contains more ground-truth information of illumination and colors, which however, has been underestimated when learning to cleanse the shadow region for most existing studies. 
In this section, we further propose a novel intra-convolution distillation loss to enhance the recovery quality of the shadow region by injecting the information of the non-shadow outputs.
Our motive comes from the fact that the boundary pixels provide color transition between the shadow region and the non-shadow region. Thus, the boundary information can be well utilized to guide the recovery of the shadow region.
\begin{figure}
    \centering
    \includegraphics[width=0.9\linewidth]{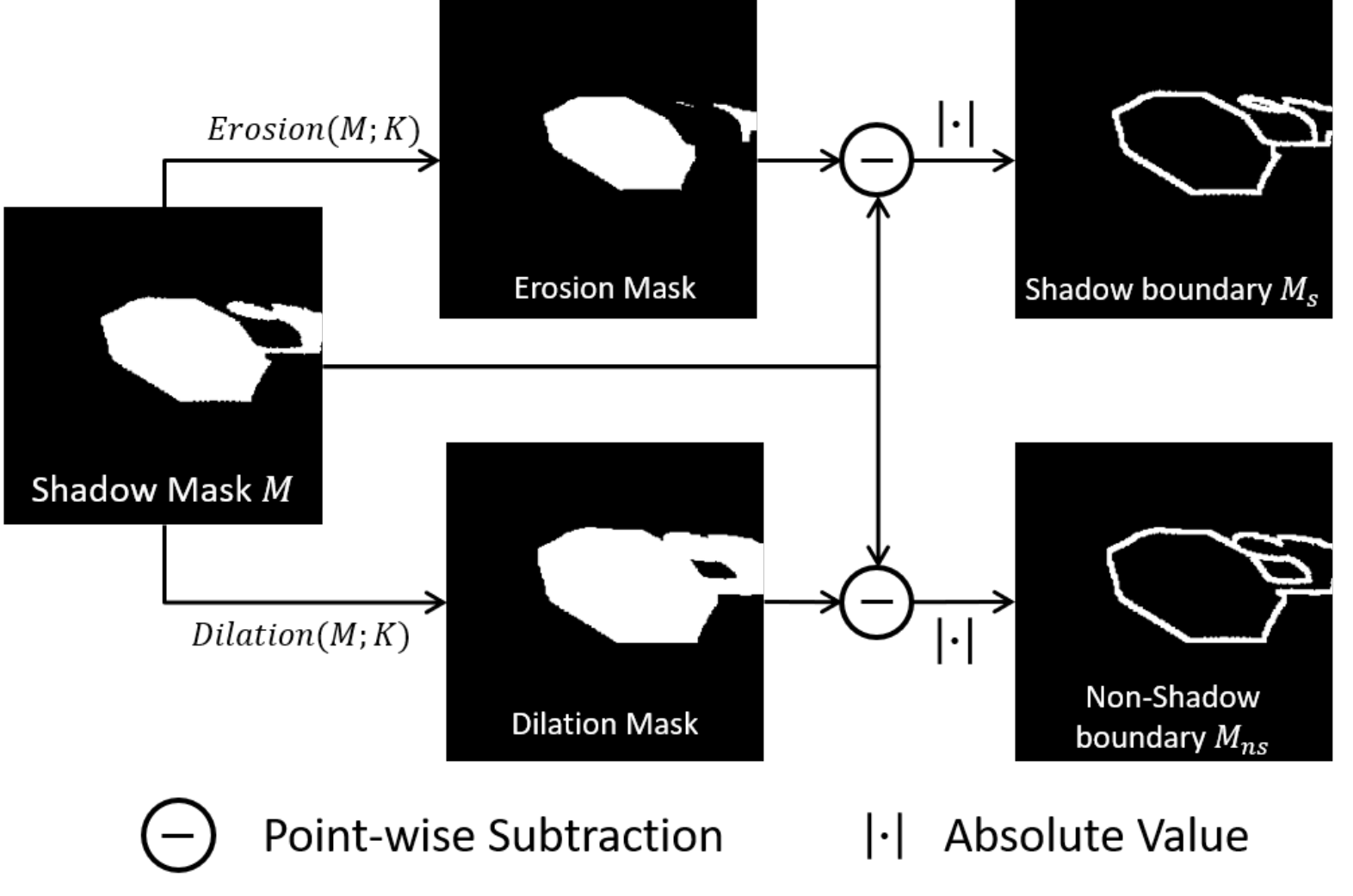}
    \caption{The shadow dilation and erosion module.}
    \label{fig:mask_manipulation}
\end{figure}

As shown in Fig.\,\ref{fig:mask_manipulation}, we respectively erode and dilate the shadow mask to obtain the boundary masks of the shadow area and the non-shadow area as:
\begin{align}
    M_{s} = M - Erosion(M; K), \\
    M_{ns} = Dilation(M; K) - M,
\end{align}
where $Erosion(\cdot \, ; K)$ and $Dilation(\cdot \, ; K)$ represent morphology erosion and dilation with a kernel $K$ of an adjustable size $\kappa \times \kappa$~\cite{haralick1987image}. 
The $M_s$ represents the boundary mask within the shadow region while $M_{ns}$ denotes the boundary mask within the non-shadow region.
Therefore, the boundary pixels in convolutional results of shadow region can be represented as ${X_{out}^s}_{|M_s=1}$ while these of non-shadow region is ${X_{out}^{ns}}_{|M_{ns}=1}$. Also, their channel size is $C$.

We aim to align the color distributions of the shadow and non-shadow boundaries in each SADC. The mean pixel value along the spatial domain is a simple-yet-effective way to represent the channel-wise color distribution of a certain image region.
To this end, we first model the color distribution as:
%

\begin{align}
    & p = softmax\left(\frac{\sum_{h=1}^H\sum_{w=1}^W{{(X_{out}^{ns}}_{|M_{ns}=1})}_{:,h,w}}{H\times W}\right), \\
    & q = softmax\left(\frac{\sum_{h=1}^H\sum_{w=1}^W{{(X_{out}^s}_{|M_{s}=1})}_{:,h,w}}{H\times W}\right).
\end{align}

Recall that the non-shadow region has more ground-truth color information, thus the non-shadow distribution $p$ provides rich supervision to enhance the color consistency in the recovery of the shadow region. Therefore, we further propose to regulate the distribution of color information in the shadow boundary given that in the non-shadow boundary region. And accordingly, an intra-convolution distillation loss is formulated as:
\begin{equation}
    \mathcal{L}_{intra} = KL(p, q),
\end{equation}
where $KL(\cdot, \cdot)$ represents the Kullback-Leibler divergence loss.


\subsection{Overall Objective}\label{overall}
Given that our network structure is built upon DHAN~\cite{cun2020towards}, in addition to the introduced intra-convolution distillation, we also consider the perceptual loss and gradient loss in DHAN.

The perceptual loss aims to preserve the structure of the image with a pretrained VGG feature extractor $VGG(\cdot)$ and is defined as:
\begin{align}
    \mathcal{L}_{percep} = ||VGG(I_{gt}), VGG(I_{out})||_1.
\end{align}
%

The gradient loss measures the gradients difference between the output images and ground-truth images:
\begin{align}
    \mathcal{L}_{grad} = ||\nabla I_{gt}, \nabla I_{out}||_1,
    \label{equation:grad_loss}
\end{align}
where $\nabla$ returns the image gradient.

Thus, the overall loss for our shadow removal can be summarized as:
\begin{align}
    \mathcal{L} = \mathcal{L}_{percep} + {\lambda}_1 \cdot \mathcal{L}_{grad} + {\lambda}_2 \cdot \sum_{i=1}^{N}\mathcal{L}^i_{intra},
\end{align}
where $\lambda_1$ and $\lambda_2$ are two coefficients, $i$ denotes the intra-convolution distillation loss from the $i$-th SADC and $N$ is the number of SADC in the DHAN backbone. We experimentally find that the final performance is insensitive to $\lambda_1$ thus we simply set $\lambda_1 = 1$ in our implementation.
In contrast, $\lambda_2$ is not a constant since the non-shadow information becomes more accurate as the training proceeds and the $\lambda_2$ needs to grow larger to inject more information to the shadow recovery module in the proposed SADC.
In the first few warmup epochs, we set $\lambda_2 = 0$ to enable the convolution for the non-shadow region to learn some basic color transformations. Then, the coefficient $\lambda_2$ increases gradually, and reaches $1$ in the end.

\section{Experiments}\label{sec:experiment}

\begin{table*}[htbp]
\caption{Quantitative shadow removal results of our networks compared with other state-of-the-art shadow removal methods on the SRD dataset. $\uparrow$ denotes the higher the metric is, the better better performance of the methods in the corresponding region, and vice versa.}
\label{Table:Quant_res_istd}
\centering
\resizebox{\linewidth}{!}{
\begin{tabular}{cll|ccc|ccc|ccc}
\hline
\multicolumn{3}{c|}{\multirow{2}{*}{Methods}} & \multicolumn{3}{c|}{Shadow Region (S)} & \multicolumn{3}{c|}{Non-Shadow Region (NS)} & \multicolumn{3}{c}{All image (ALL)} \\ \cline{4-12} 
\multicolumn{3}{c|}{} & RMSE $\downarrow$ & PSNR $\uparrow$ & SSIM $\uparrow$ & RMSE $\downarrow$ & PSNR $\uparrow$ & SSIM $\uparrow$ & RMSE $\downarrow$ & PSNR $\uparrow$ & SSIM $\uparrow$ \\ \hline
\multicolumn{3}{c|}{Input Image} &32.10  &22.40  &0.9361  &7.09  &27.32  &0.9755  &10.88  &20.56  &0.8934  \\ \hline
\multicolumn{3}{c|}{Guo~\emph{et al.}~\cite{guo2012paired}} &17.44  &29.19  &0.9547  &15.13  &21.69  &0.7538  &15.48  &20.51  &0.6997  \\
\multicolumn{3}{c|}{Gong~\emph{et al.}~\cite{gong2016interactive}} &13.54  &30.14  &0.9727  &7.20  &26.98  &\underline{0.9730}  &8.03  &24.71  &0.9360  \\
\multicolumn{3}{c|}{ST-CGAN~\cite{wang2018stacked}} &14.09  &32.60  &0.9604  &12.87  &25.01  &0.7580  &13.06  &23.95  &0.7080  \\
\multicolumn{3}{c|}{Param+M+D-Net~\cite{le2019shadow}} &11.84  &31.43  &0.9811  &7.51  &26.21  &0.9687  &7.94  &24.69  &0.9413  \\
\multicolumn{3}{c|}{DSC~\cite{hu2019direction}} &8.72  &34.65  &0.9835  &5.04  &\underline{31.26}  &0.9691  &5.59  &29.00  &0.9438  \\
\multicolumn{3}{c|}{SP+M-Net~\cite{le2020shadow}} &9.64  &32.89  &0.9861  &7.73  &26.11  &0.9650  &7.96  &25.01  &0.9483  \\
\multicolumn{3}{c|}{DHAN\cite{cun2020towards}} &7.73  &35.53  &0.9882  &5.29  &31.05  &0.9705  &5.66  &29.11  &0.9543  \\
\multicolumn{3}{c|}{DHAN\dag} &7.34  &36.08  &0.9901  &5.65  &30.30  &0.9659  &5.90  &28.64  &0.9533  \\
\multicolumn{3}{c|}{Auto-Exposure~\cite{fu2021auto}} &7.91  &34.71  &0.9752  &5.51  &28.61  &0.8799  &5.89  &27.19  &0.8456  \\ \hline
\multicolumn{3}{c|}{Ours + Detection Mask} & \underline{6.64}  & \underline{36.68} &\underline{0.9907}  &\underline{4.97}  &30.81  &0.9721  & \underline{5.22}  & \underline{29.21}  & \underline{0.9595}  \\
\multicolumn{3}{c|}{Ours} & \textbf{6.00}  & \textbf{38.07} &\textbf{0.9918}  &\textbf{4.48}  &\textbf{32.30}  &\textbf{0.9779}  & \textbf{4.70}  & \textbf{30.63}  & \textbf{0.9675}  \\ \hline
\end{tabular}
}
\end{table*}

\begin{figure*}[htbp]
\centering
\subfigure[\tiny{Shadow Image}]{
\begin{minipage}[t]{0.09\textwidth}
\centering
\includegraphics[width=\linewidth, height=\linewidth]{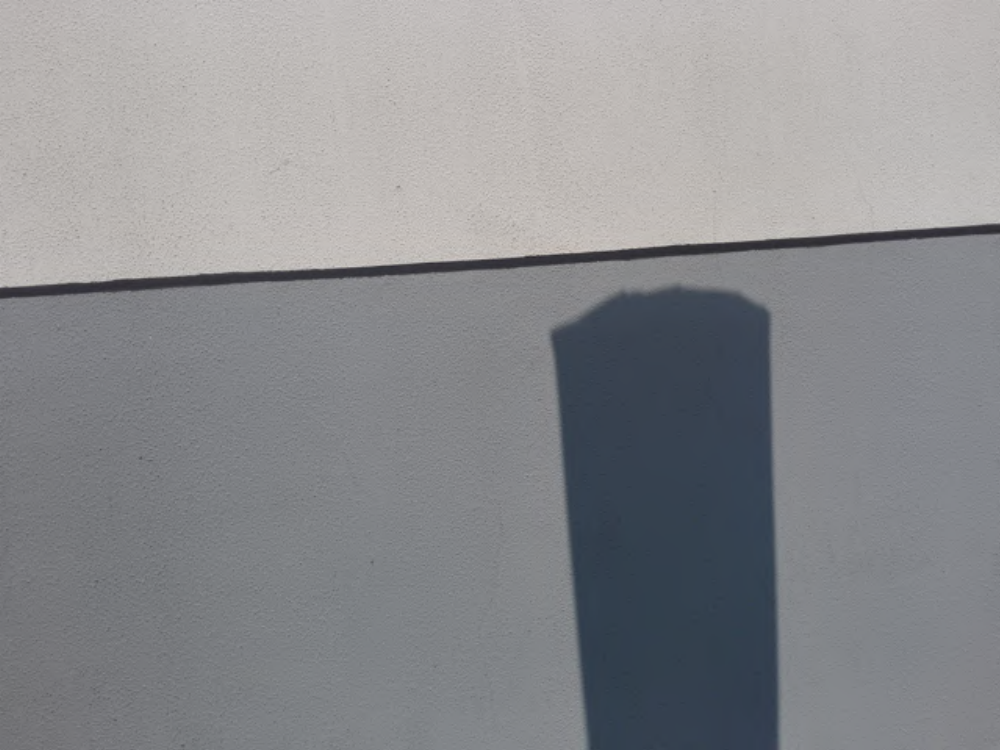} \\
\vspace{0.05cm}
\includegraphics[width=\linewidth, height=\linewidth]{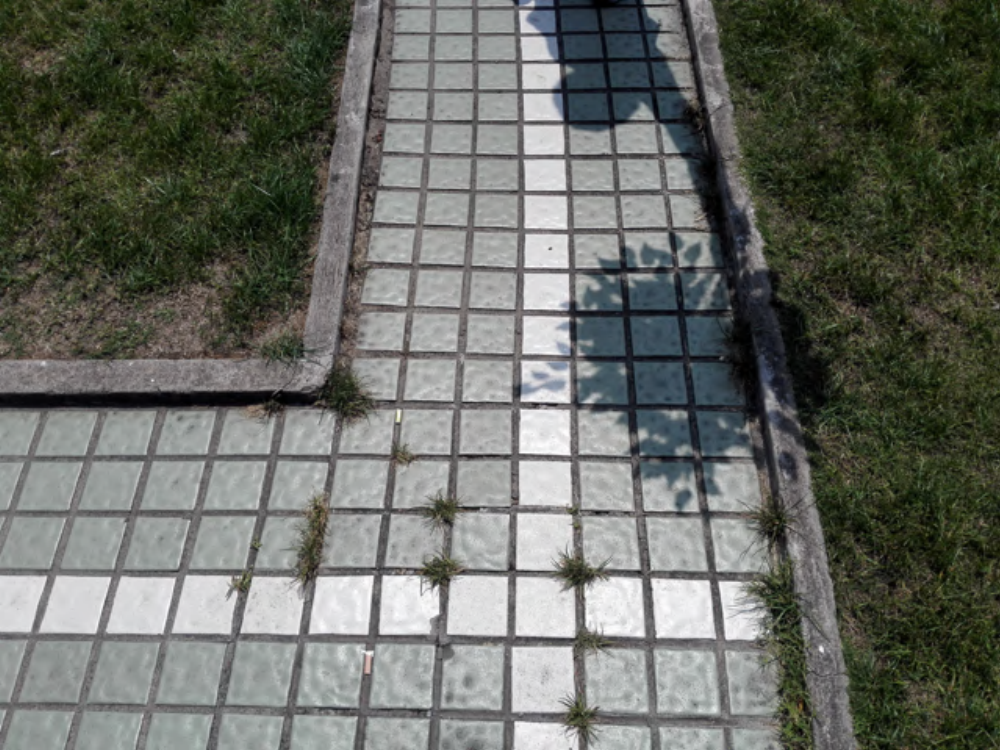} \\
\vspace{0.05cm}
\includegraphics[width=\linewidth, height=\linewidth]{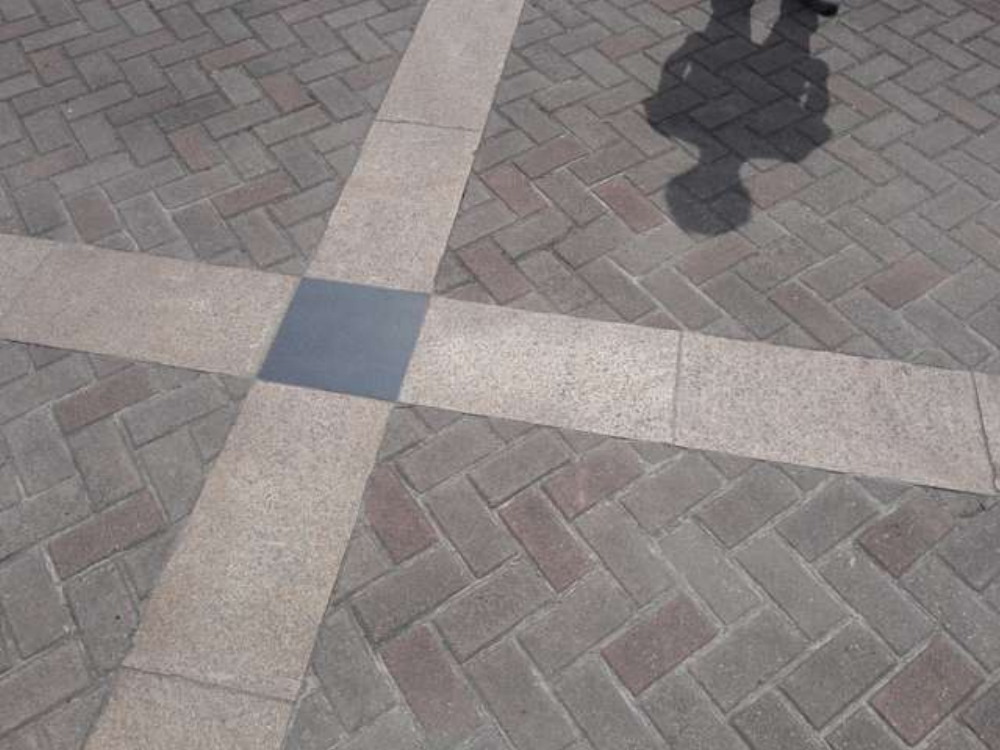} \\
\vspace{0.05cm}
\includegraphics[width=\linewidth, height=\linewidth]{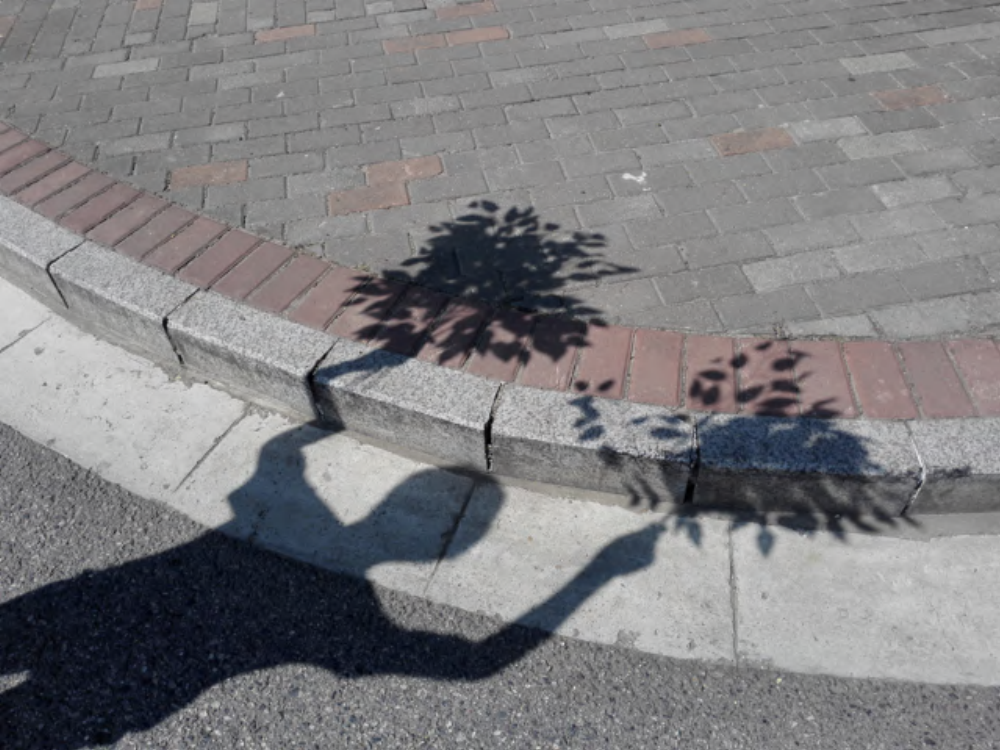}
\end{minipage}}
\hspace{-0.5em}
\subfigure[\tiny{Guo \emph{et al.}}]{
\begin{minipage}[t]{0.09\textwidth}
\centering
\includegraphics[width=\linewidth, height=\linewidth]{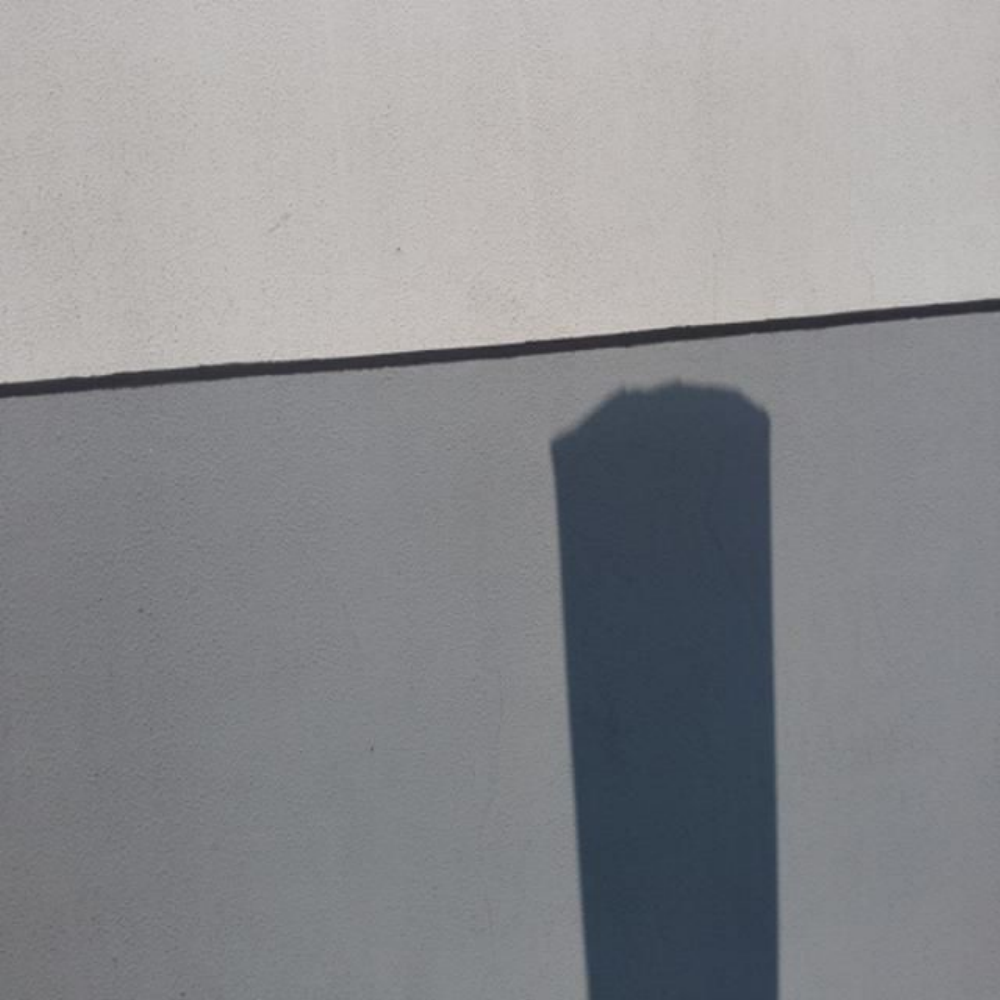} \\
\vspace{0.05cm}
\includegraphics[width=\linewidth, height=\linewidth]{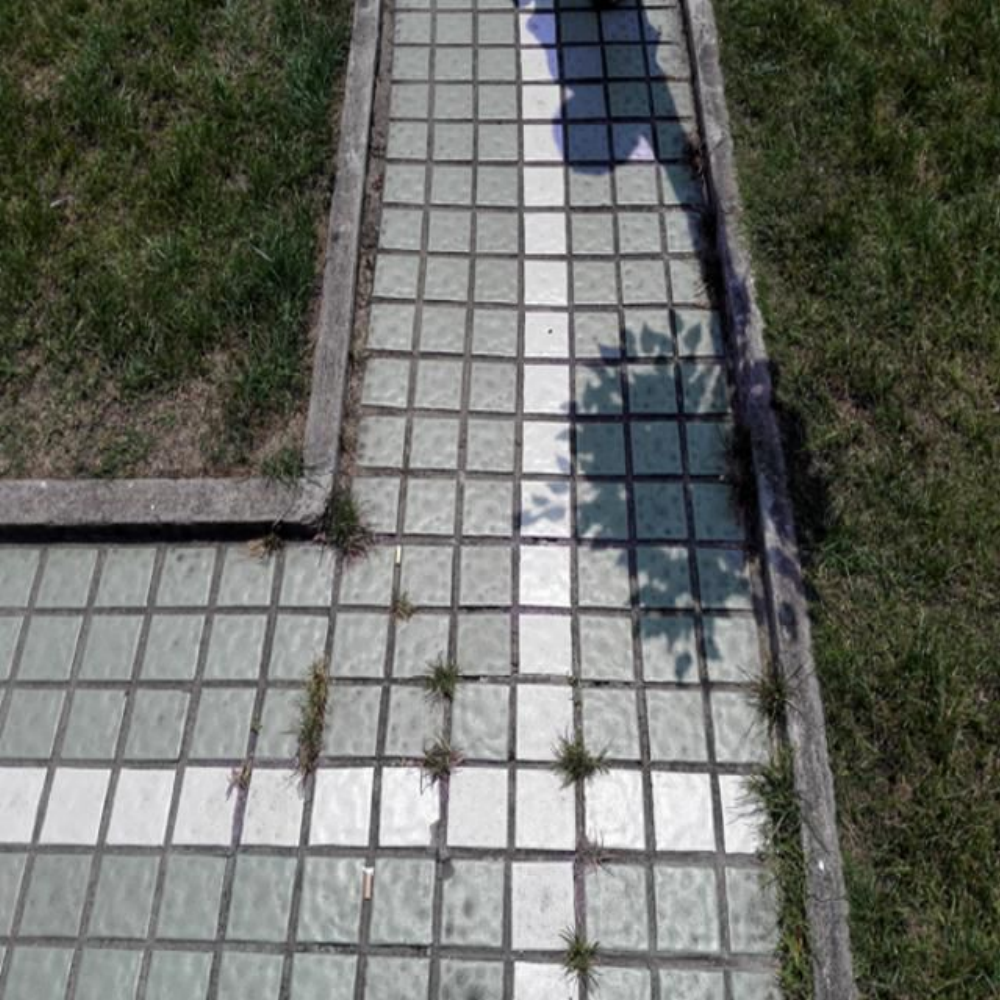} \\
\vspace{0.05cm}
\includegraphics[width=\linewidth, height=\linewidth]{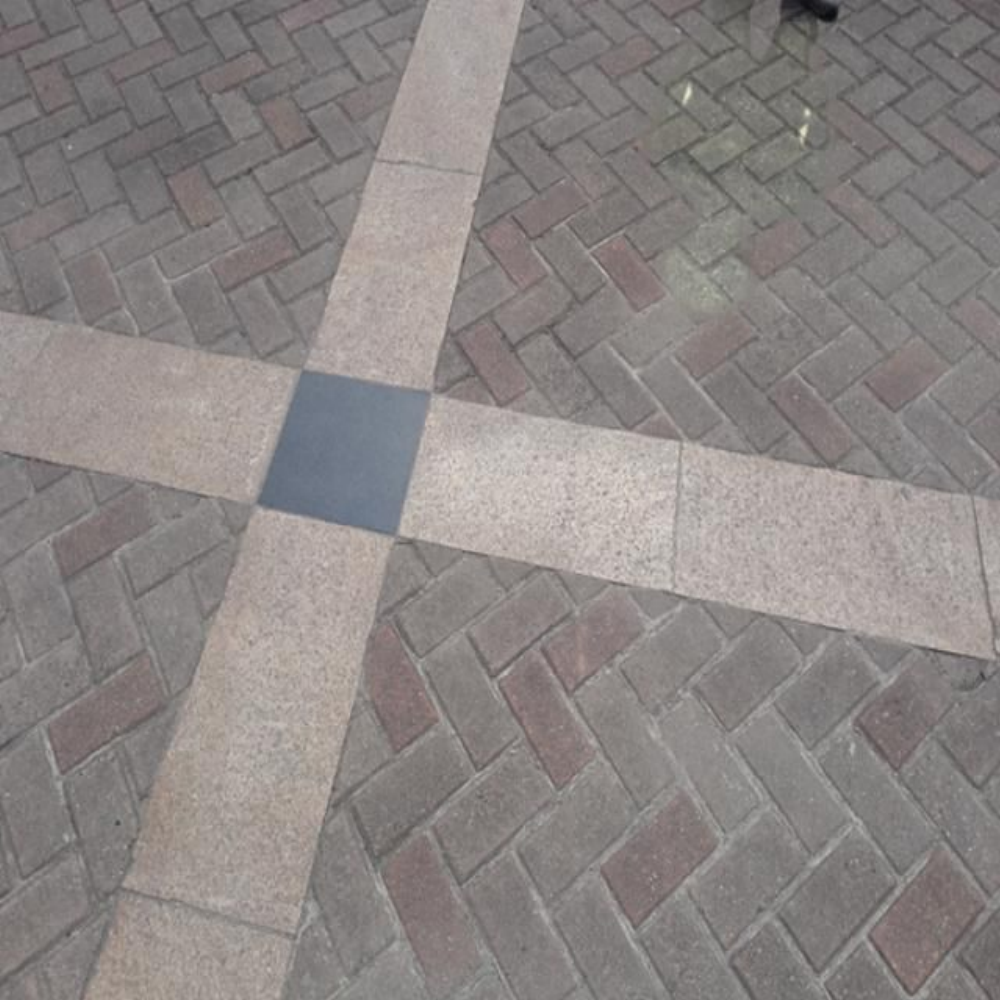} \\
\vspace{0.05cm}
\includegraphics[width=\linewidth, height=\linewidth]{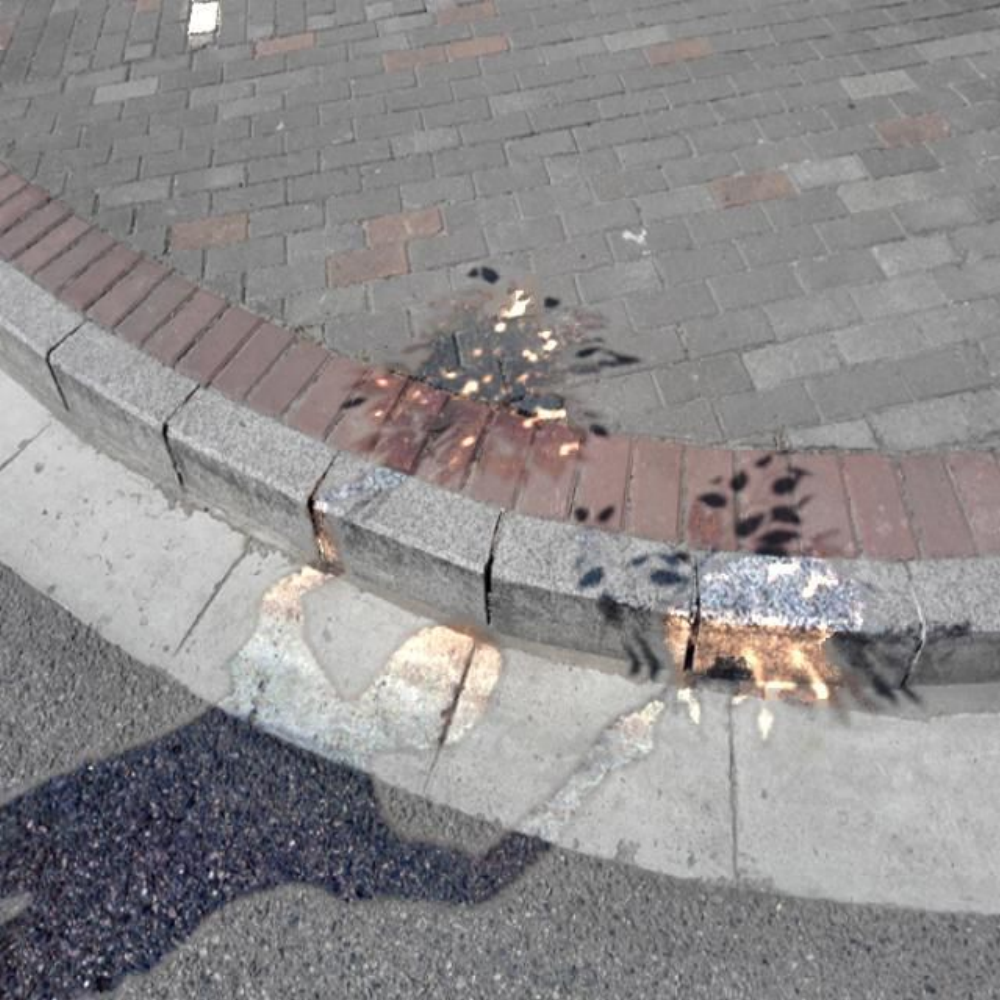}
\end{minipage}}
\hspace{-0.5em}
\subfigure[\tiny{ST-CGAN}]{
\begin{minipage}[t]{0.09\textwidth}
\centering
\includegraphics[width=\linewidth, height=\linewidth]{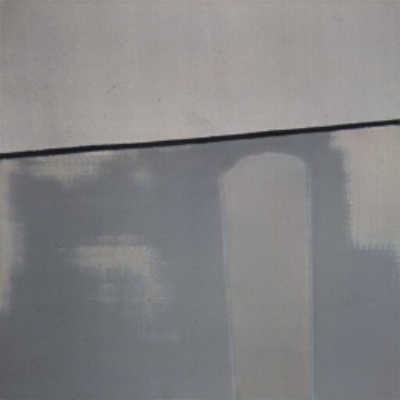} \\
\vspace{0.05cm}
\includegraphics[width=\linewidth, height=\linewidth]{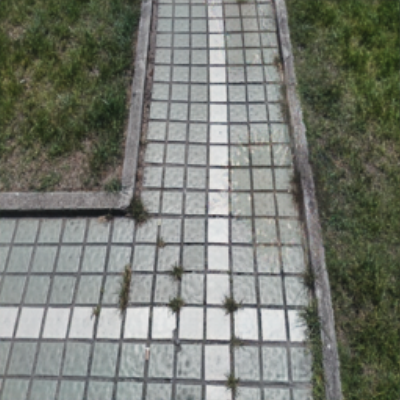} \\
\vspace{0.05cm}
\includegraphics[width=\linewidth, height=\linewidth]{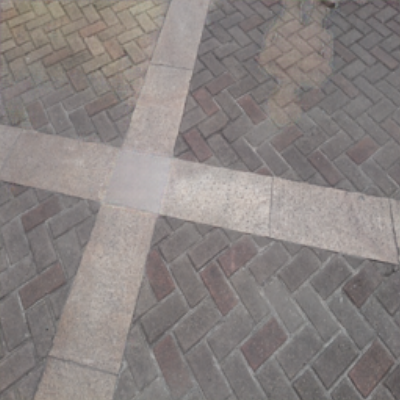} \\
\vspace{0.05cm}
\includegraphics[width=\linewidth, height=\linewidth]{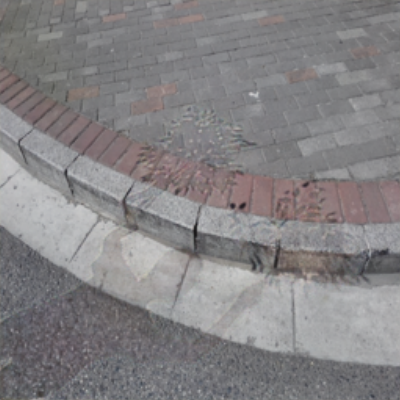}
\end{minipage}}
\hspace{-0.5em}
\subfigure[\tiny{\tiny{Param+M+D-Net}}]{
\begin{minipage}[t]{0.09\textwidth}
\centering
\includegraphics[width=\linewidth, height=\linewidth]{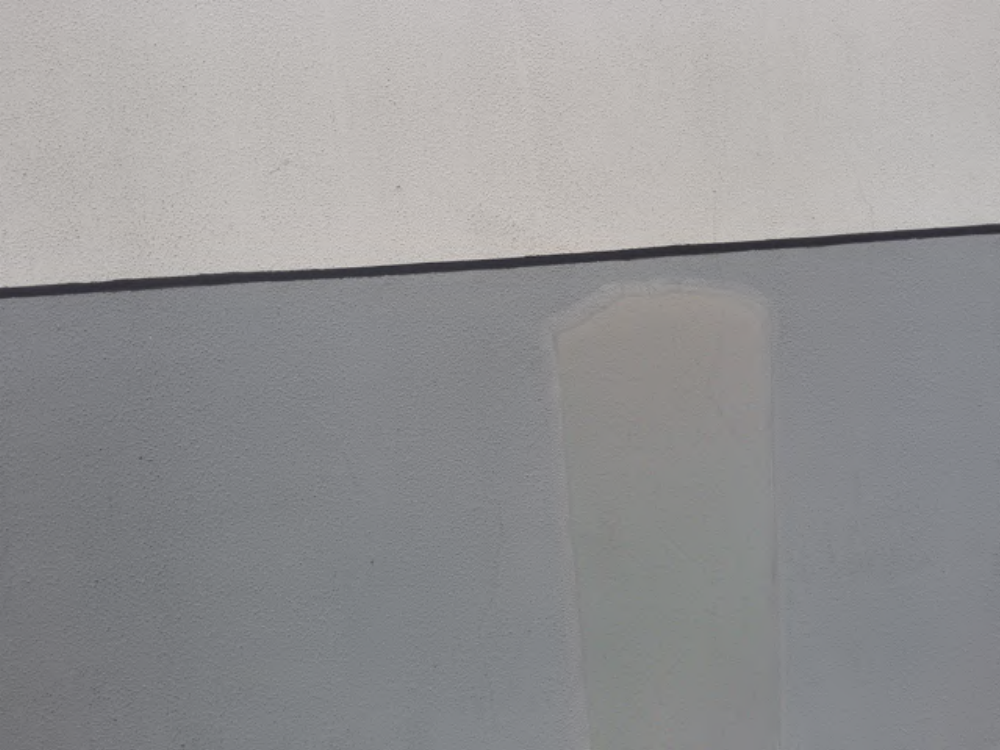} \\
\vspace{0.05cm}
\includegraphics[width=\linewidth, height=\linewidth]{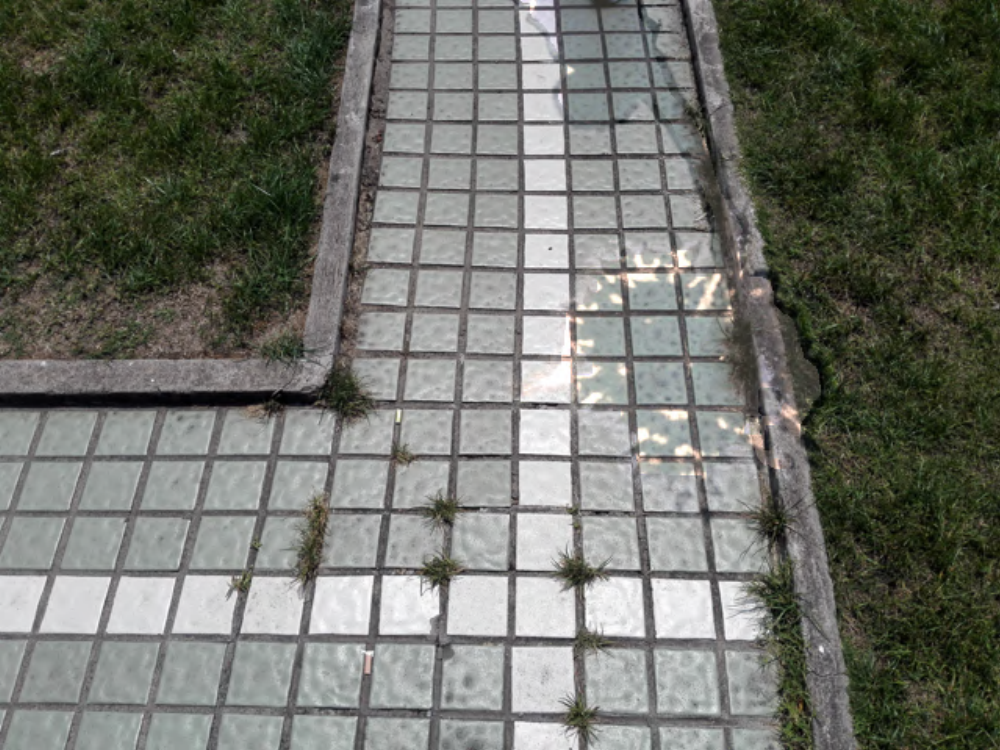} \\
\vspace{0.05cm}
\includegraphics[width=\linewidth, height=\linewidth]{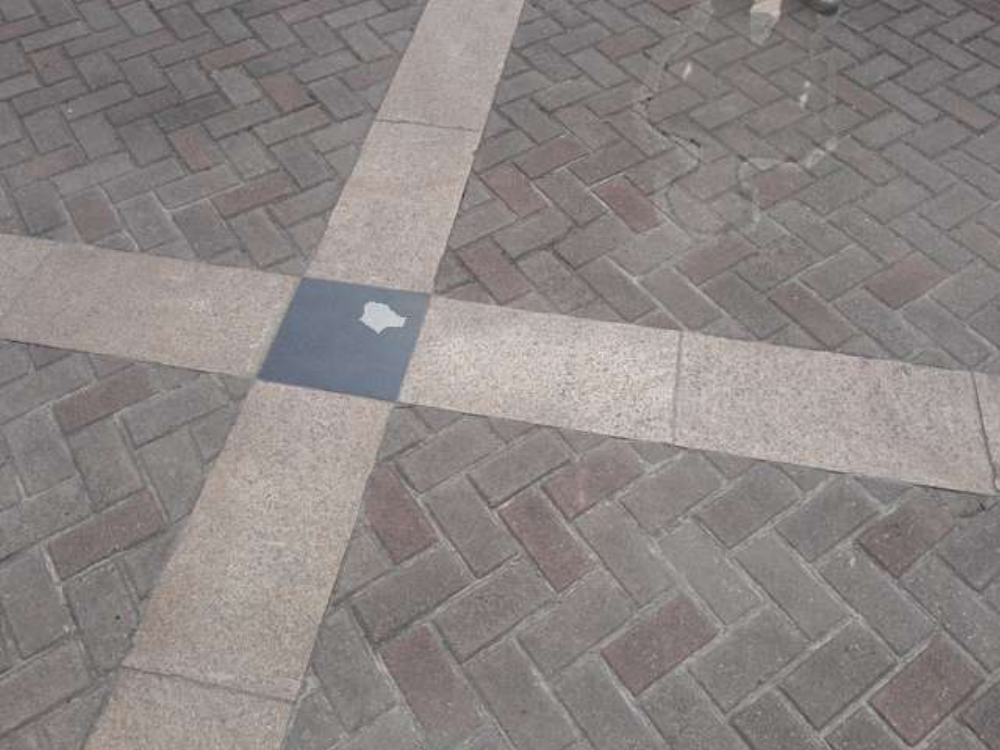} \\
\vspace{0.05cm}
\includegraphics[width=\linewidth, height=\linewidth]{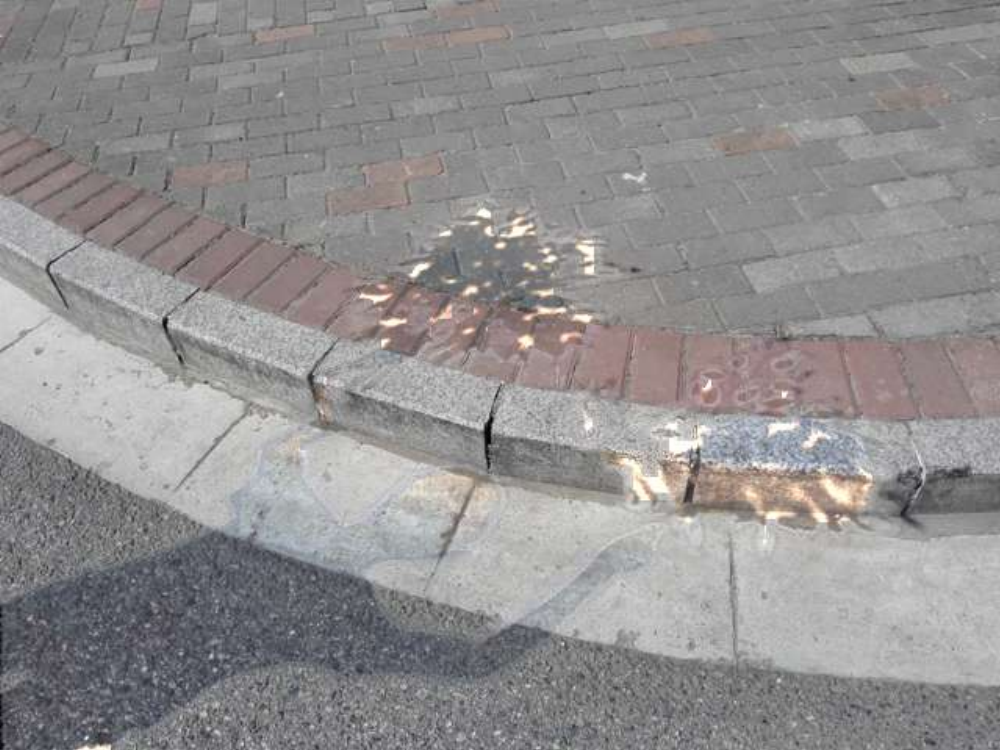}
\end{minipage}}
\hspace{-0.5em}
\subfigure[\tiny{DSC}]{
\begin{minipage}[t]{0.09\textwidth}
\centering
\includegraphics[width=\linewidth, height=\linewidth]{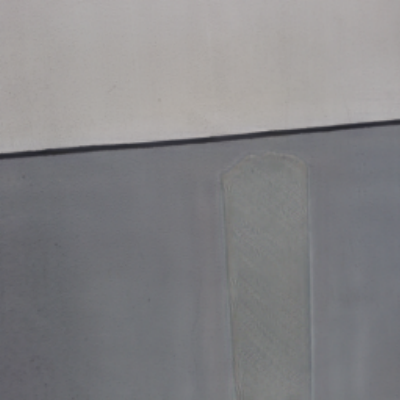} \\
\vspace{0.05cm}
\includegraphics[width=\linewidth, height=\linewidth]{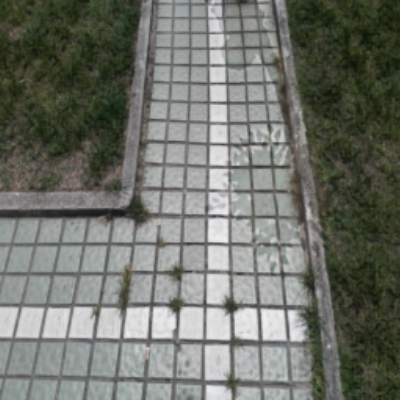} \\
\vspace{0.05cm}
\includegraphics[width=\linewidth, height=\linewidth]{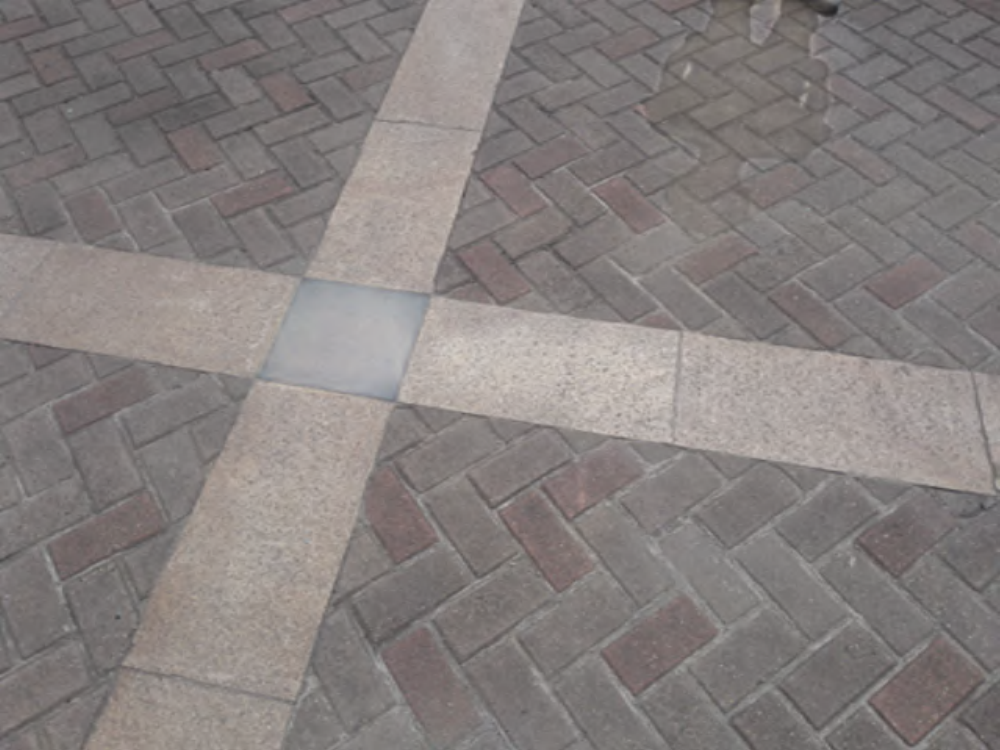} \\
\vspace{0.05cm}
\includegraphics[width=\linewidth, height=\linewidth]{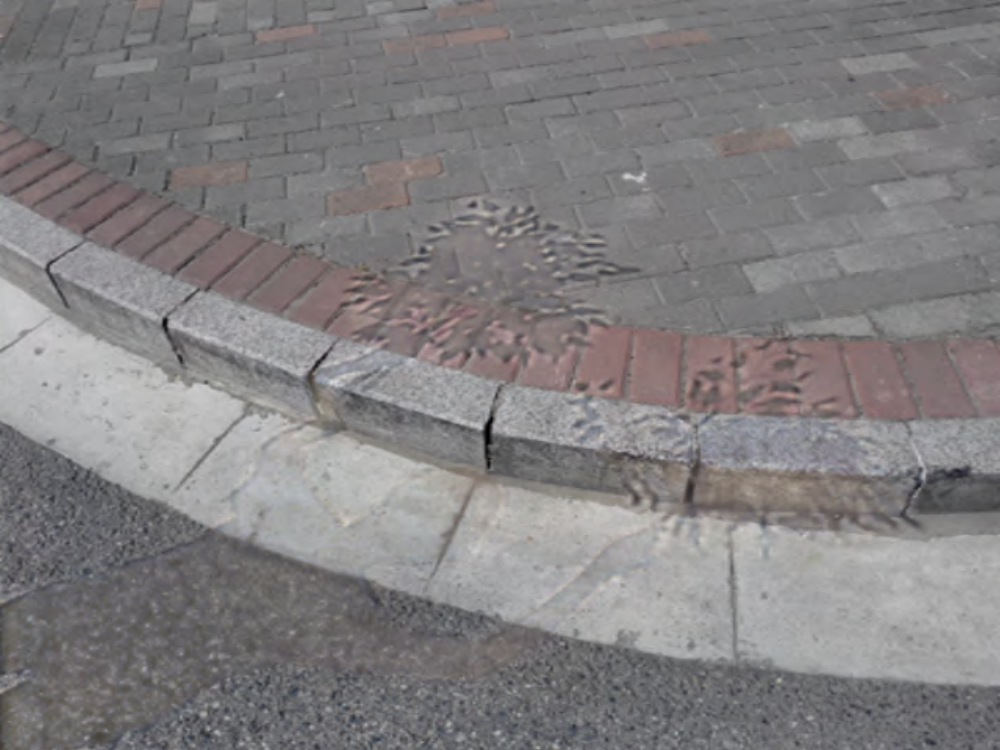}
\end{minipage}}
\hspace{-0.5em}
\subfigure[\tiny{SP+M-Net}]{
\begin{minipage}[t]{0.09\textwidth}
\centering
\includegraphics[width=\linewidth, height=\linewidth]{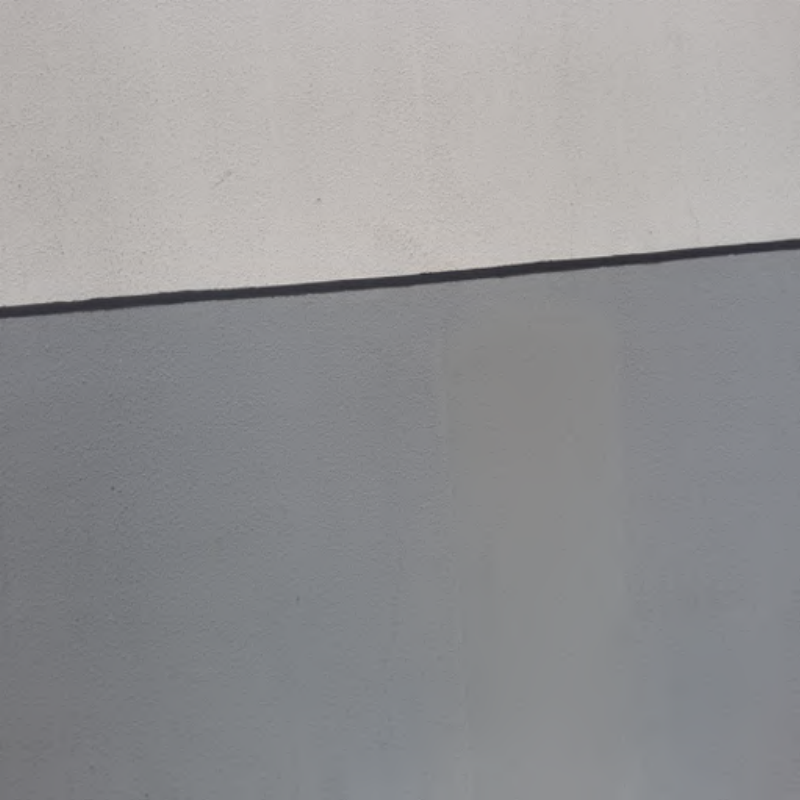} \\
\vspace{0.05cm}
\includegraphics[width=\linewidth, height=\linewidth]{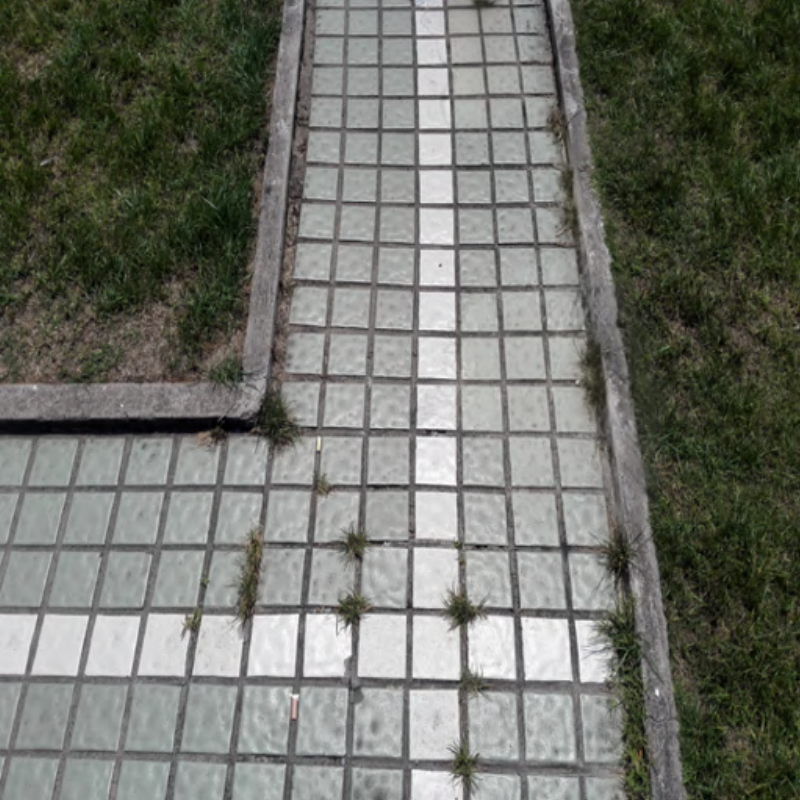} \\
\vspace{0.05cm}
\includegraphics[width=\linewidth, height=\linewidth]{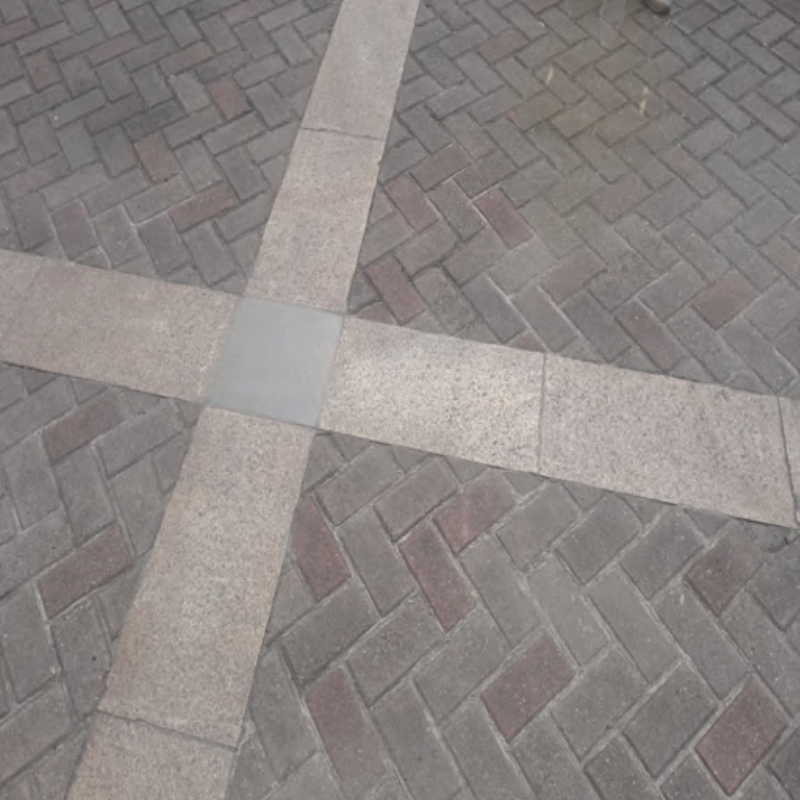} \\
\vspace{0.05cm}
\includegraphics[width=\linewidth, height=\linewidth]{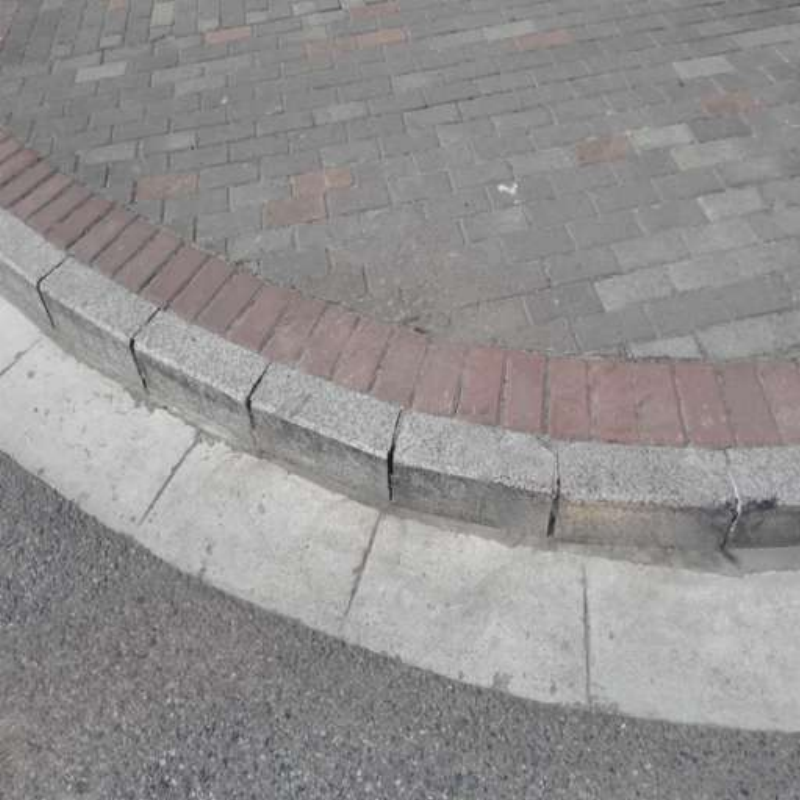}
\end{minipage}}
\hspace{-0.5em}
\subfigure[\tiny{DHAN}]{
\begin{minipage}[t]{0.09\textwidth}
\centering
\includegraphics[width=\linewidth, height=\linewidth]{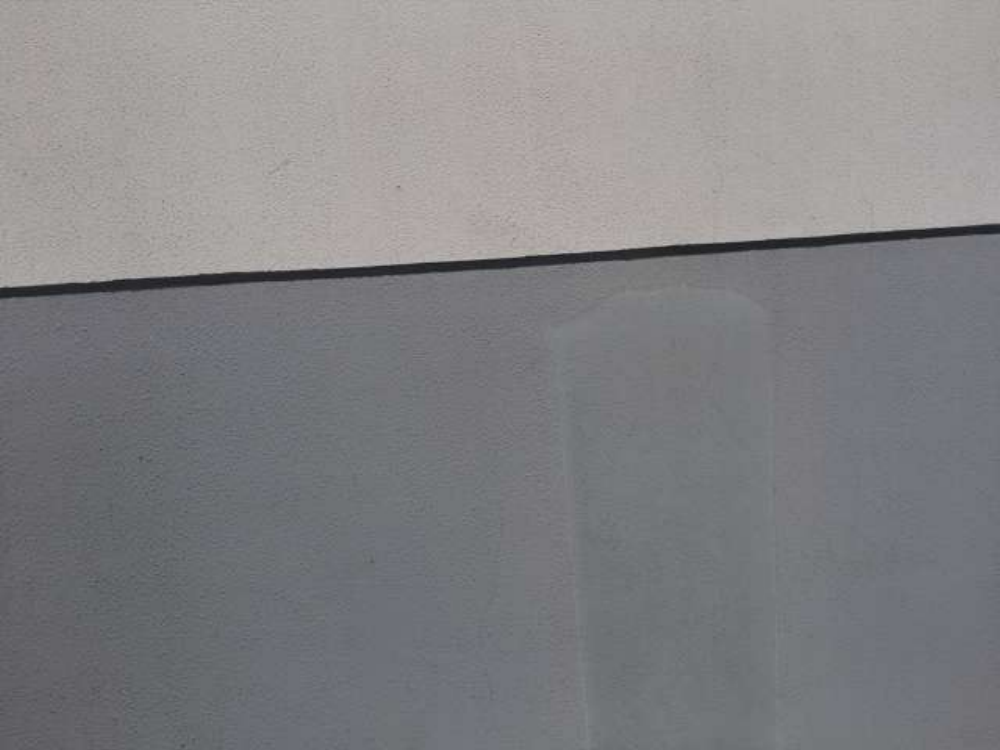} \\
\vspace{0.05cm}
\includegraphics[width=\linewidth, height=\linewidth]{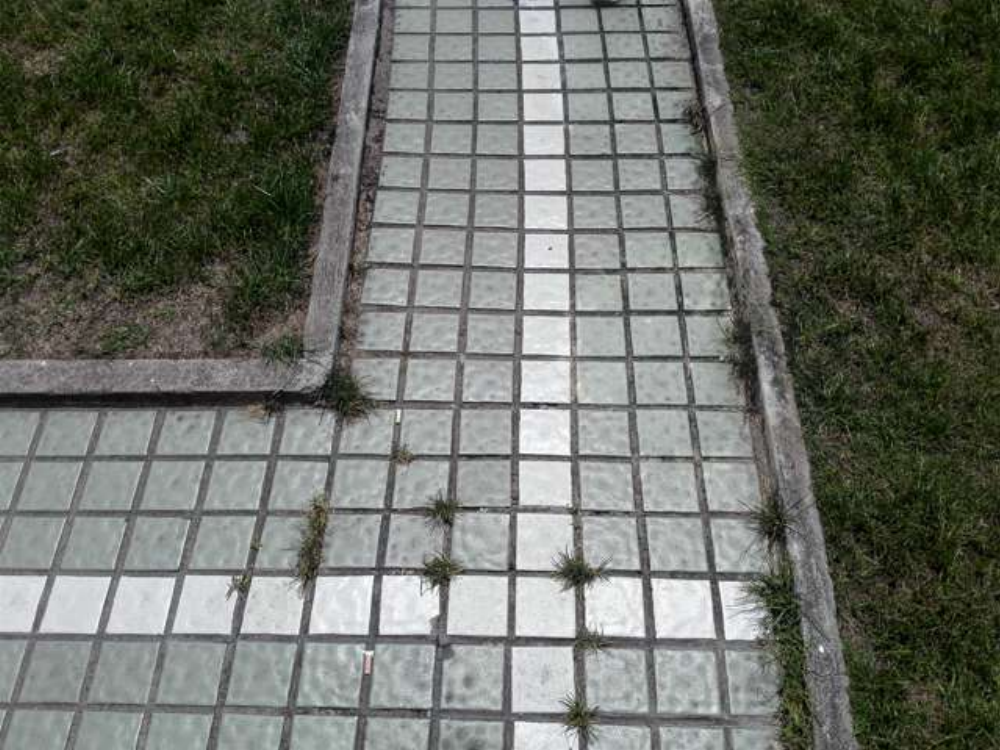} \\
\vspace{0.05cm}
\includegraphics[width=\linewidth, height=\linewidth]{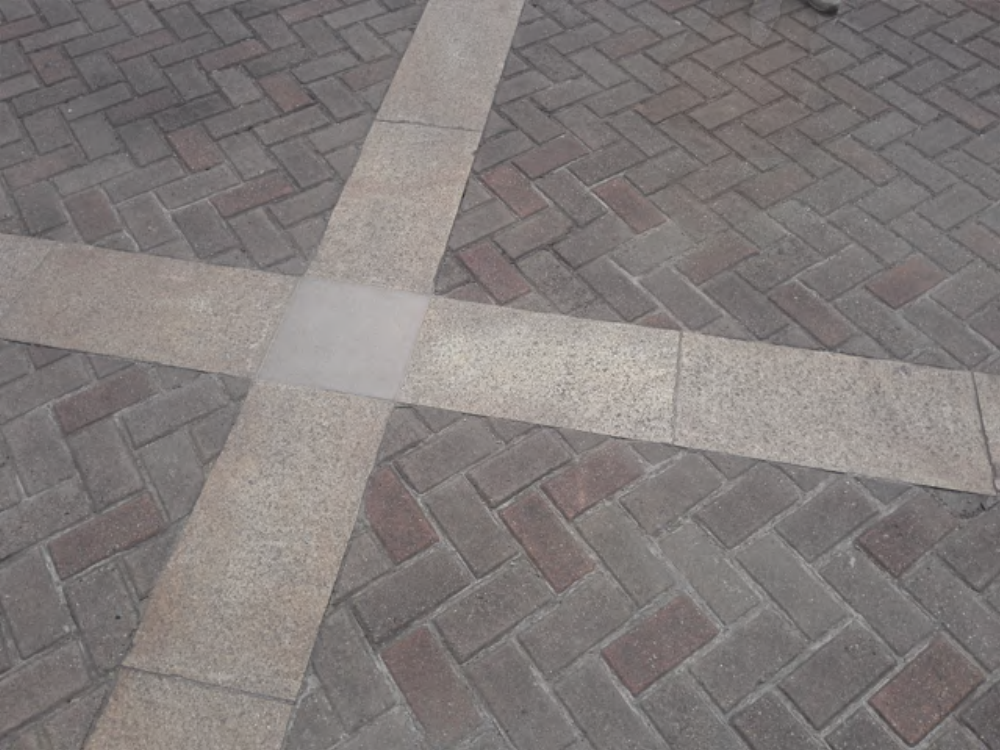} \\
\vspace{0.05cm}
\includegraphics[width=\linewidth, height=\linewidth]{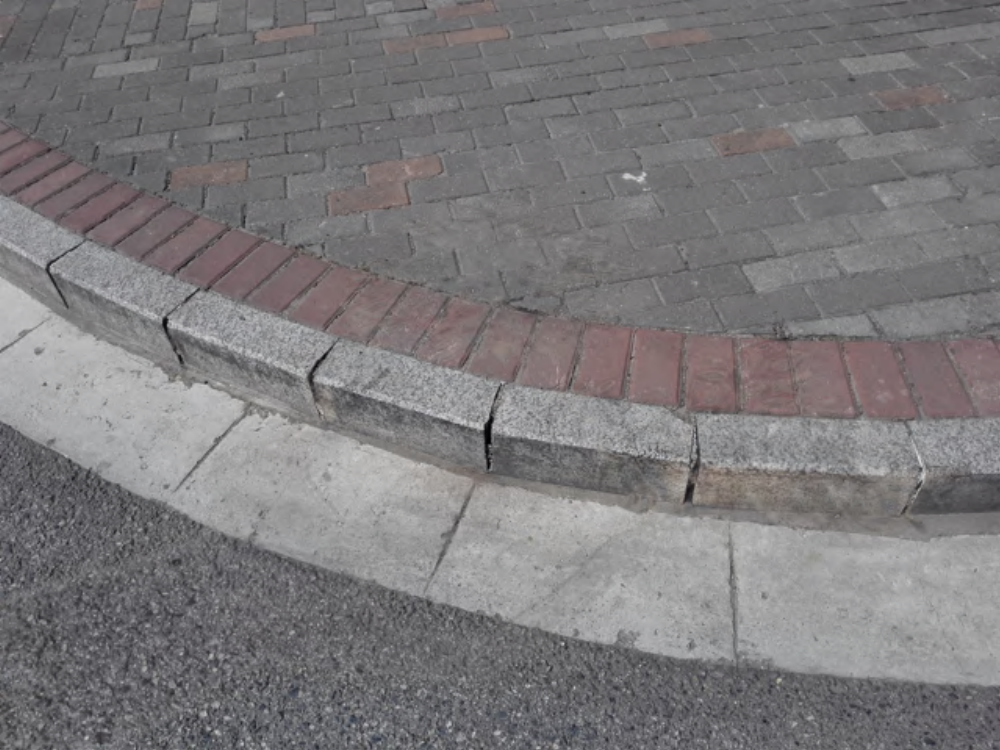}
\end{minipage}}
\hspace{-0.5em}
\subfigure[\tiny{Auto-Exposure}]{
\begin{minipage}[t]{0.09\textwidth}
\centering
\includegraphics[width=\linewidth, height=\linewidth]{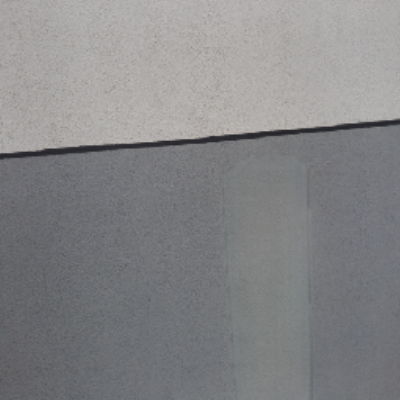} \\
\vspace{0.05cm}
\includegraphics[width=\linewidth, height=\linewidth]{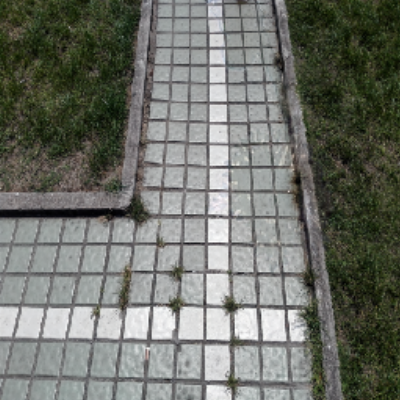} \\
\vspace{0.05cm}
\includegraphics[width=\linewidth, height=\linewidth]{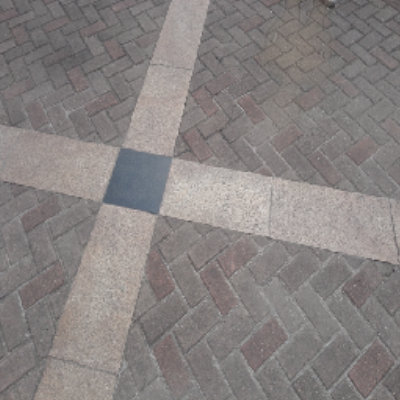} \\
\vspace{0.05cm}
\includegraphics[width=\linewidth, height=\linewidth]{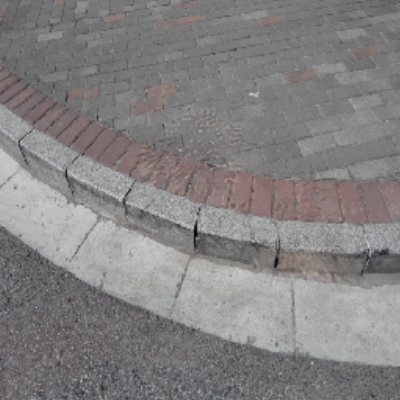}
\end{minipage}}
\hspace{-0.5em}
\subfigure[\tiny{Ours}]{
\begin{minipage}[t]{0.09\textwidth}
\centering
\includegraphics[width=\linewidth, height=\linewidth]{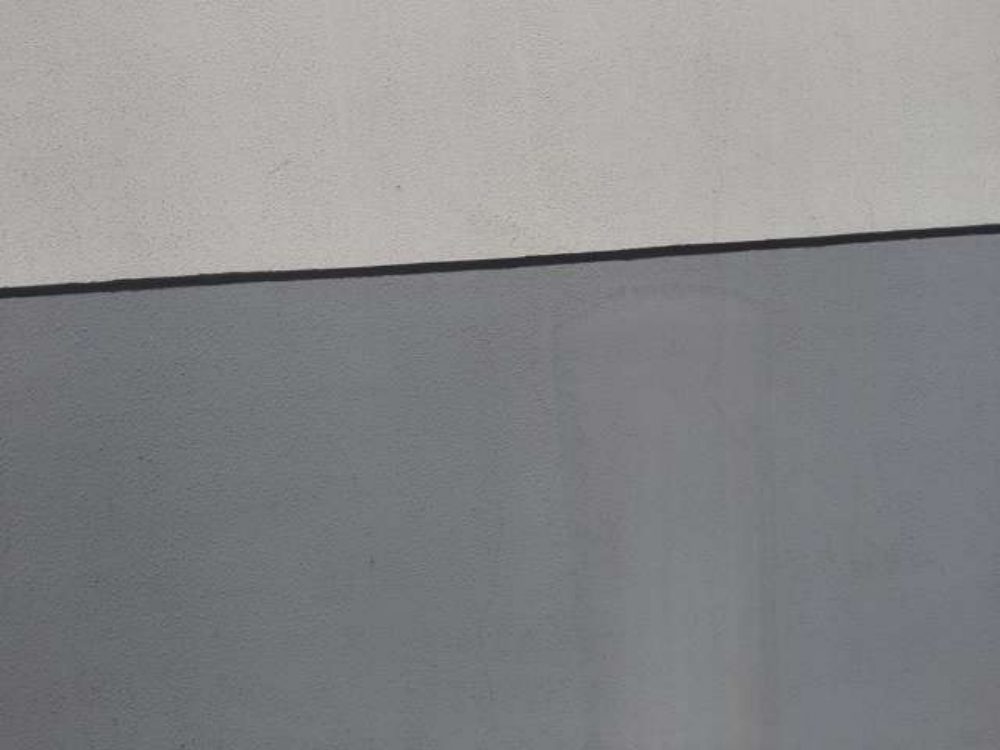} \\
\vspace{0.05cm}
\includegraphics[width=\linewidth, height=\linewidth]{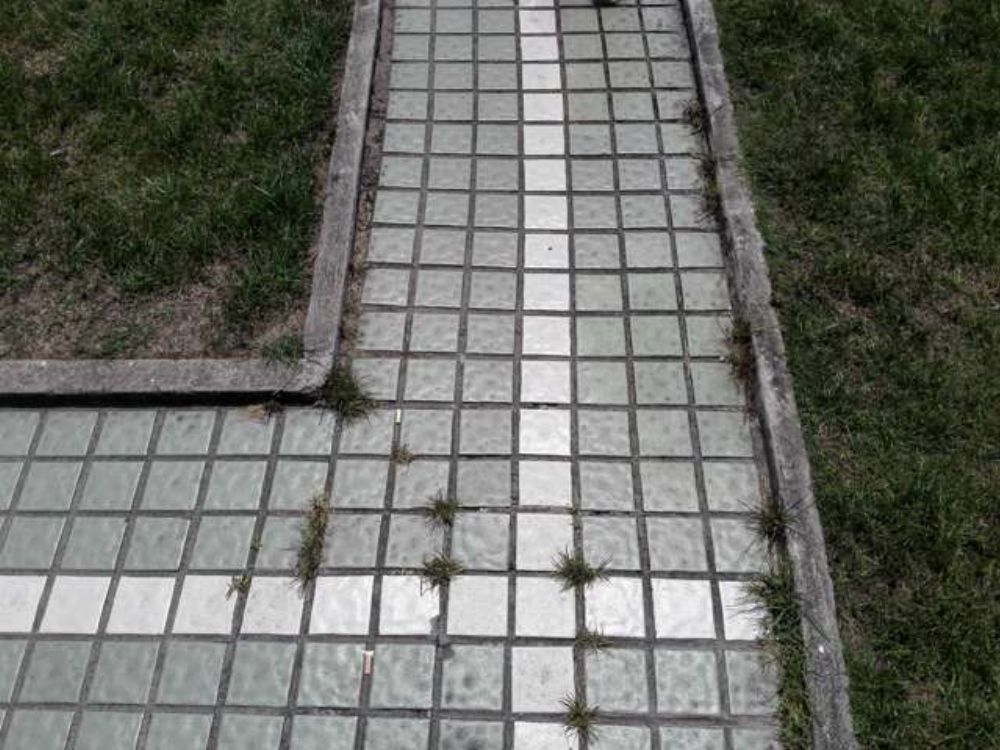} \\
\vspace{0.05cm}
\includegraphics[width=\linewidth, height=\linewidth]{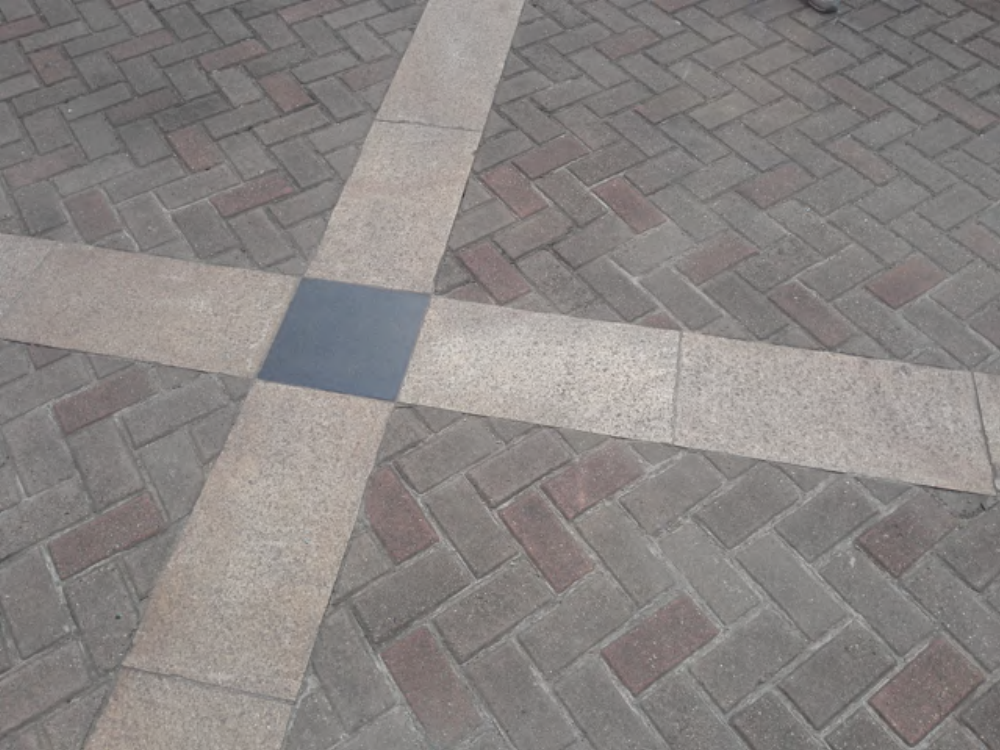} \\
\vspace{0.05cm}
\includegraphics[width=\linewidth, height=\linewidth]{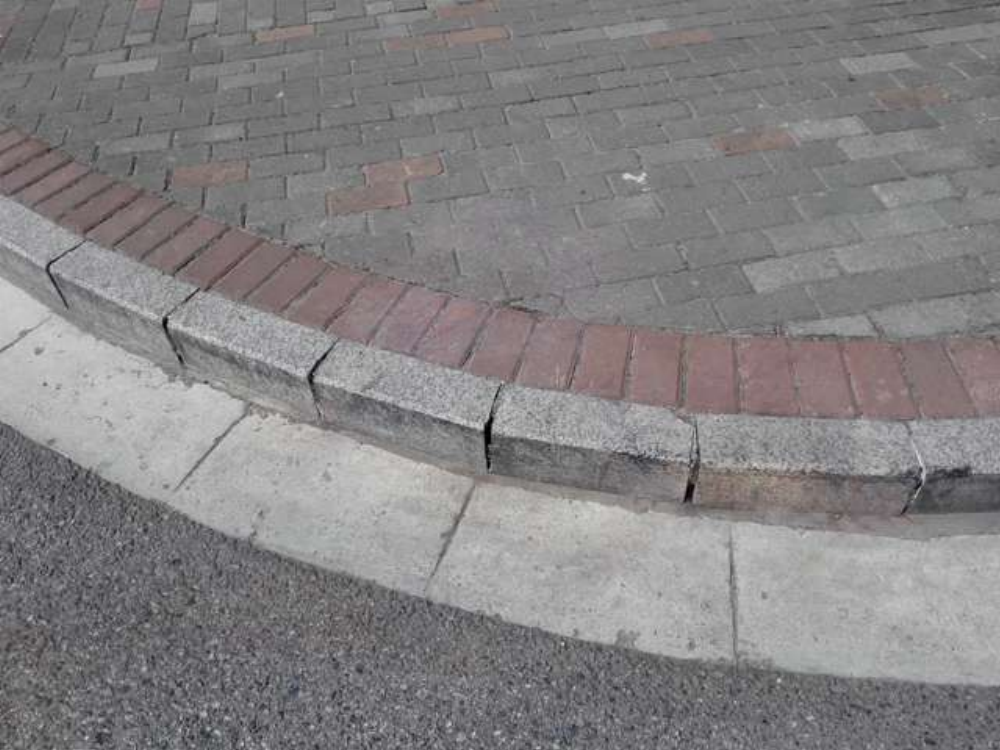}
\end{minipage}}
\hspace{-0.5em}
\subfigure[\tiny{GT}]{
\begin{minipage}[t]{0.09\textwidth}
\centering
\includegraphics[width=\linewidth, height=\linewidth]{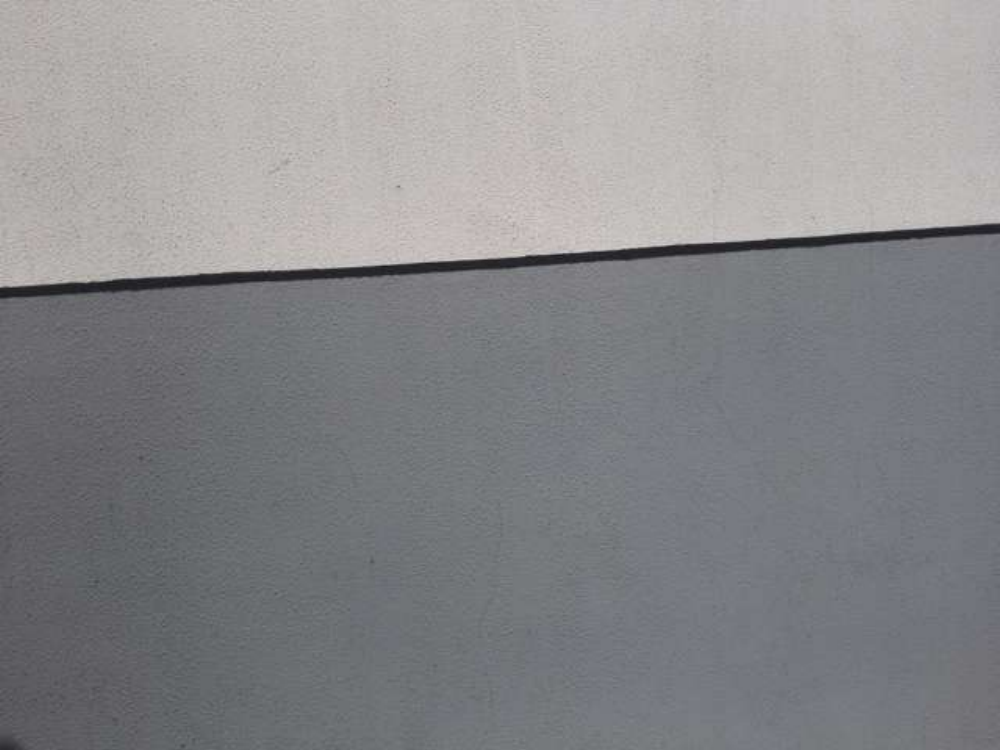} \\
\vspace{0.05cm}
\includegraphics[width=\linewidth, height=\linewidth]{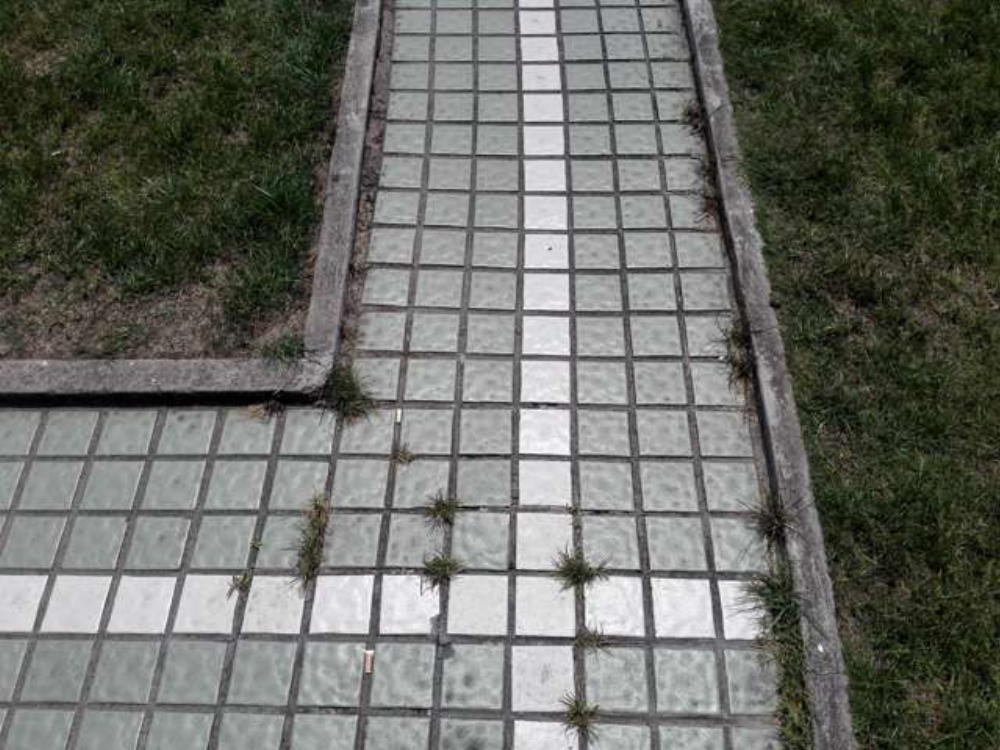} \\
\vspace{0.05cm}
\includegraphics[width=\linewidth, height=\linewidth]{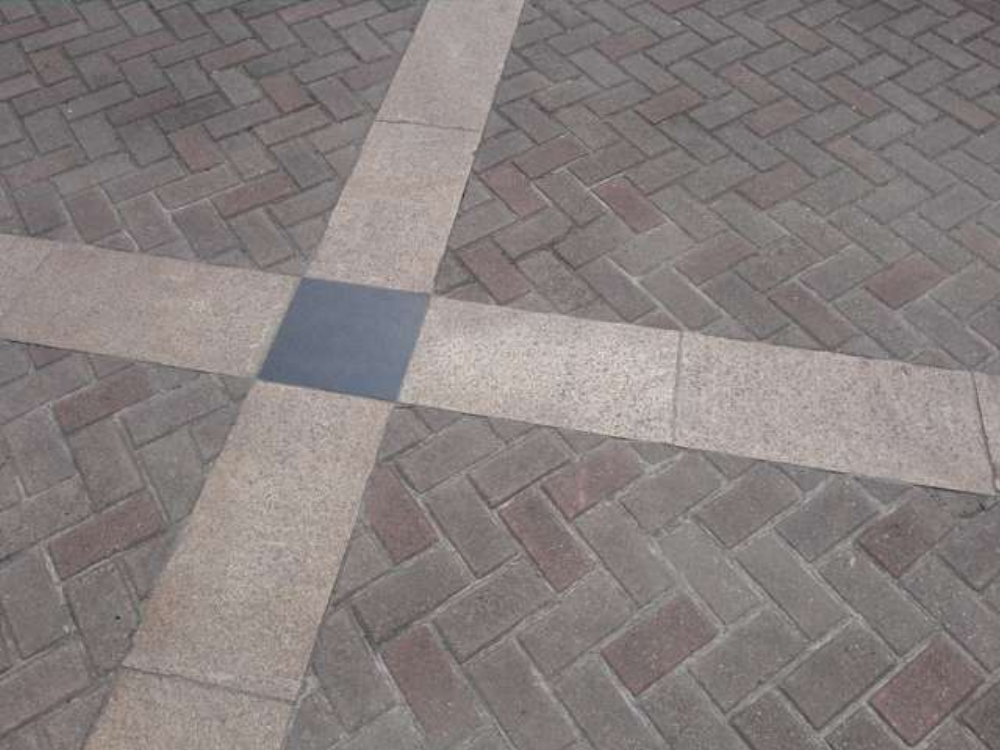} \\
\vspace{0.05cm}
\includegraphics[width=\linewidth, height=\linewidth]{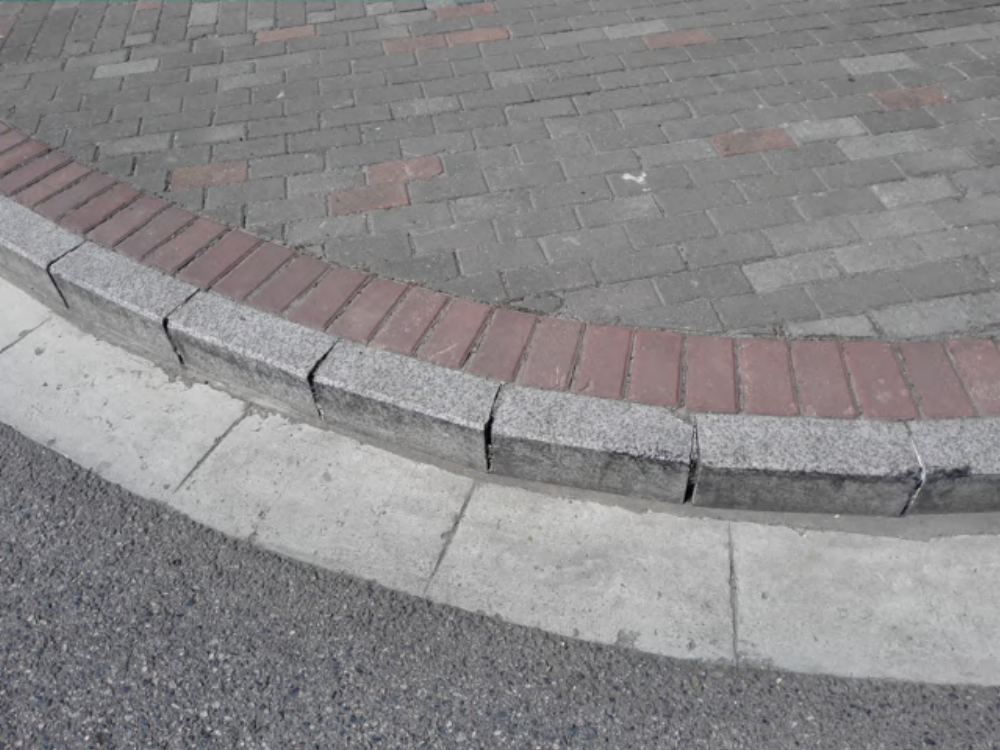}
\end{minipage}}
\caption{Visualization of shadow removal results of SADC and current methods including Guo \emph{et al.}~\cite{guo2012paired}, ST-CGAN~\cite{wang2018stacked}, Param+M-D-Net~\cite{le2019shadow}, DSC~\cite{hu2019direction}, SP+M-Net~\cite{le2020shadow}, DHAN~\cite{cun2020towards}and Auto-Exposure~\cite{fu2021auto}, on ISTD dataset.}
\label{fig:istd_vis}
\end{figure*}

\subsection{Datasets and Evaluation Metrics}\label{dataset}

\textbf{Datasets}. We train and evaluate the proposed SADC on two public datasets including ISTD~\cite{wang2018stacked} and SRD~\cite{qu2017deshadownet}. The ISTD dataset contains 1,330 training triplets (shadow image, shadow mask, and shadow-free image) from 134 scenes, whereas the testing set consists of 540 triplets from 45 scenes. The SRD dataset contains 2680 pairs (shadow image and shadow-free image) of training samples whose shadow masks are unknown and the testing set has 408 pairs of shadow images. We utilize the public shadow mask provided by DHAN~\cite{cun2020towards} for both training and evaluation.

\textbf{Evaluation Metrics}. We use the root mean square error (RMSE) in the LAB color space between the shadow removal results and their corresponding ground-truth images to evaluate the model efficacy. We also report the peak signal-to-noise ratio (PSNR) and structural similarity (SSIM) to evaluate the shadow recovery results of various methods. We compare the performance with previous works~\cite{guo2012paired,gong2016interactive,wang2018stacked,le2019shadow,le2020shadow,hu2019direction,cun2020towards,fu2021auto}. 

\textbf{Implementation Details}. %
We implement our method using PyTorch and test it on a single A100 GPU.
The DHAN without shadow detection branch is denoted as DHAN\dag to tell from the original DHAN~\cite{cun2020towards}. We replace the original convolutions in the main block of DHAN\dag with our plug-and-play SADC module to demonstrate the efficacy of our method.
For the ISTD dataset, the input size of images is $256 \times 256$ and the batch size is set to 5. We utilize 300 epochs to train the whole network. 
As for the SRD dataset, since the SRD dataset contains larger input samples and is more difficult to converge, the input size of images is set to $400 \times 400$ and the batch size is set to 2. 
The training epoch is set to 500 to reach a better convergence. 
For both datasets, the Adam optimizer with the learning rate 0.0002 is adopted. The dilation ratio $\kappa$ is set to 7, and the warmup epoch for intra-class distillation loss is set to 50.

\begin{table*}[htp]
\centering
\caption{Quantitative shadow removal results of our networks compared with other state-of-the-art shadow removal methods on the SRD dataset. $\uparrow$ denotes the higher the metric is, the better better performance of the methods in the corresponding region, and vice versa.}
\resizebox{\linewidth}{!}{
\begin{tabular}{cll|ccc|ccc|ccc}
\hline
\multicolumn{3}{c|}{\multirow{2}{*}{Methods}} & \multicolumn{3}{c|}{Shadow Region (S)} & \multicolumn{3}{c|}{Non-Shadow Region (NS)} & \multicolumn{3}{c}{All image (ALL)} \\ \cline{4-12} 
\multicolumn{3}{c|}{} & RMSE $\downarrow$ & PSNR $\uparrow$ & SSIM $\uparrow$ & RMSE $\downarrow$ & PSNR $\uparrow$ & SSIM $\uparrow$ & RMSE $\downarrow$ & PSNR $\uparrow$ & SSIM $\uparrow$ \\ \hline
\multicolumn{3}{c|}{Input Image} &36.62   &18.98   &0.8720   &4.54   &31.68   &0.9815   &13.83   &18.26   &0.8372   \\ \hline
\multicolumn{3}{c|}{DSC~\cite{hu2019direction}} &19.35  &26.46  &0.9033  &15.76  &24.80  &0.7740  &16.77  &21.64  &0.6566  \\
\multicolumn{3}{c|}{DHAN~\cite{cun2020towards}} &\underline{6.94}  &\underline{33.84}  &\underline{0.9797}  &\textbf{3.63}  &\underline{35.06}  &\textbf{0.9850}  &\underline{4.61}  &\underline{30.74}  &\textbf{0.9577}  \\
\multicolumn{3}{c|}{DHAN\dag} &7.96  &33.13  &0.9740  &4.22  &33.82  &0.9748  &5.32  &29.80  &0.9398  \\
\multicolumn{3}{c|}{Auto-Exposure~\cite{fu2021auto}} &8.33  &32.44  &0.9684  &5.48  &30.84  &0.9504  &6.25  &27.97  &0.9016  \\ \hline
\multicolumn{3}{c|}{Ours} &\textbf{6.49}  &\textbf{35.69}  &\textbf{0.9823}  &\underline{3.69}  &\textbf{36.34}  &\underline{0.9827}  &\textbf{4.49}  &\textbf{32.18}  &\underline{0.9573}  \\ \hline
\end{tabular}
}
\label{Table:Quant_res_srd}
\end{table*}

\subsection{Results on ISTD dataset}

In this section, we evaluate our method on ISTD dataset quantitatively and qualitatively.
The quantitative results are presented in Table\,\ref{Table:Quant_res_istd}. We compare our method with current state-of-the-art algorithms, including Guo~\emph{et al.}~\cite{guo2012paired}, DHAN~\cite{cun2020towards}, Auto-Exposure~\cite{fu2021auto} and so on. 
%
%
Obviously, our method performs better in both shadow and non-shadow regions. Specifically, the RMSE decreases by 24.1\% and 9.8\% compared with the Auto-Exposure network in shadow and non-shadow regions, respectively. 
Furthermore, by replacing the conventional convolution with the SADC, our method outperforms DHAN\dag by reducing RMSE in the shadow region from 7.34 to 6.00 and the RMSE in the non-shadow region from 5.65 to 4.48, which demonstrates that the performance gain comes from the proposed SADC module and intra-convolution distillation loss instead of the adjusted network structure. 
%
Since the ground-truth shadow mask is often hard to obtain, we also report the shadow removal performance when no accurate shadow masks are provided in testing phases (denoted as Ours + Dectection Mask in Table\,\ref{Table:Quant_res_istd}). To be more specific, we employ the shadow mask generated by~\cite{zhu2021mitigating} (a shadow detection method) as the input of the SADC, and our method still performs better than current SOTAs on most metrics.

The visualization of our method and comparative methods are shown in Fig.\,\ref{fig:istd_vis}. Our method has the best performance in removing image shadow among all images. Specifically, our method not only recovers the shadow region effectively, but also maps the non-shadow counterpart better to its ground truth. For instance, the SP+M-Net \big((f) column\big) can also process the shadow region in the first image nicely, but it ruins the background information. The background is much lighter compared with the ground-truth image, while our method well addresses this issue.

\subsection{Results on SRD dataset}

We further evaluate the our method on SRD dataset in Table\,\ref{Table:Quant_res_srd}. Note that the SRD dataset does not have accurate shadow, making it hard to separate the shadow region and non-shadow region and thus may impair the overall performance of the proposed SADC module. However, our method still performs better \emph{w.r.t}. most metrics even under this circumstance. The RMSE in shadow region reduces from 6.94 to 6.49 compared with the DHAN and the RMSE in the non-shadow region shows a comparable result. Furthermore, the PSNR value increases significantly in both the shadow region and the non-shadow region, which means our methods can recover the shadow images with more details. Although the backbone network (\emph{i.e.}, DHAN\dag) is not able to remove shadow effectively, our method still achieves compelling shadow removal results with the help of SADC and intra-convolution dilation loss, which further demonstrates the effectiveness of our method.

\subsection{Ablation study}\label{ablation}
In this section, we conduct comprehensive ablation studies to explore the effectiveness of the proposed SADC module and intra-class distillation loss. To ensure fairness, all ablation experiments are conducted with the same training hyper-parameters on ISTD dataset, as detailed in Sec.\,\ref{dataset}.

\begin{table}[htp]
    \tabcaption{Efficacy of different components in our method.}
    \resizebox{\linewidth}{!}{
    \begin{tabular}{ccc|ccc}
    \hline
    \multicolumn{3}{c|}{Method} & \multicolumn{3}{c}{RMSE} \\ \hline
    \multicolumn{1}{c|}{baseline} & \multicolumn{1}{c|}{SADC} & \multicolumn{1}{c|}{Distillation Loss} & \multicolumn{1}{c|}{Shadow} & \multicolumn{1}{c|}{Non-shadow} & \multicolumn{1}{c}{All} \\ \hline
    \Checkmark & \multicolumn{1}{c}{} & \multicolumn{1}{c|}{} & 7.34 & \multicolumn{1}{c}{5.65} & \multicolumn{1}{c}{5.90} \\
    \multicolumn{1}{c}{\Checkmark} & \Checkmark &  & \multicolumn{1}{c}{6.17} & 4.52 & 4.77 \\
    \multicolumn{1}{c}{\Checkmark} & \Checkmark & \Checkmark & \multicolumn{1}{c}{6.00} & 4.48 & 4.70 \\ \hline
    \end{tabular}}
    \label{table:effective}
\end{table}

\begin{figure}[!t]
    \centering
    \subfigure[\tiny{Shadow Image}]{
        \begin{minipage}[ht]{0.22\linewidth}
            \includegraphics[width=\linewidth, height=\linewidth]{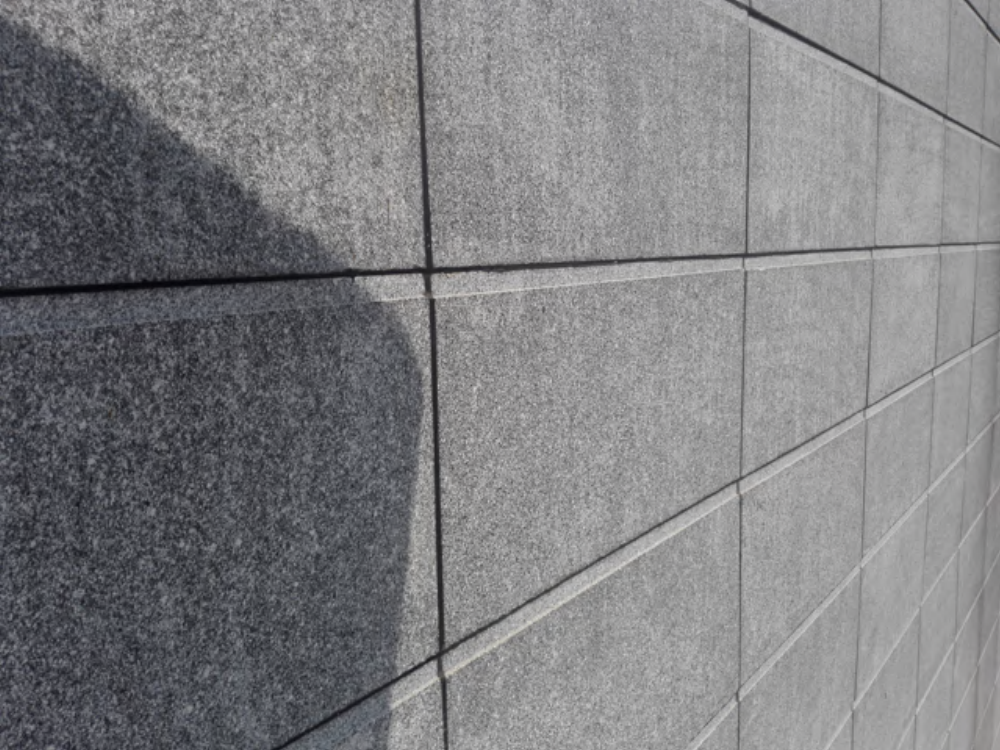} \\
            \includegraphics[width=\linewidth, height=\linewidth]{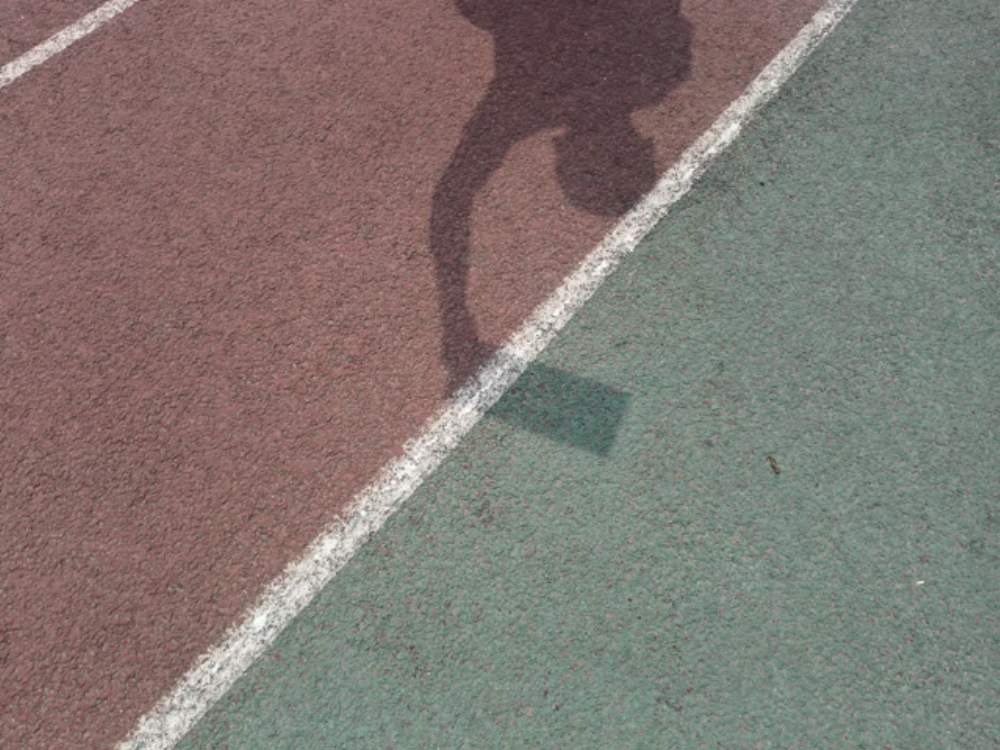} \\
            \includegraphics[width=\linewidth, height=\linewidth]{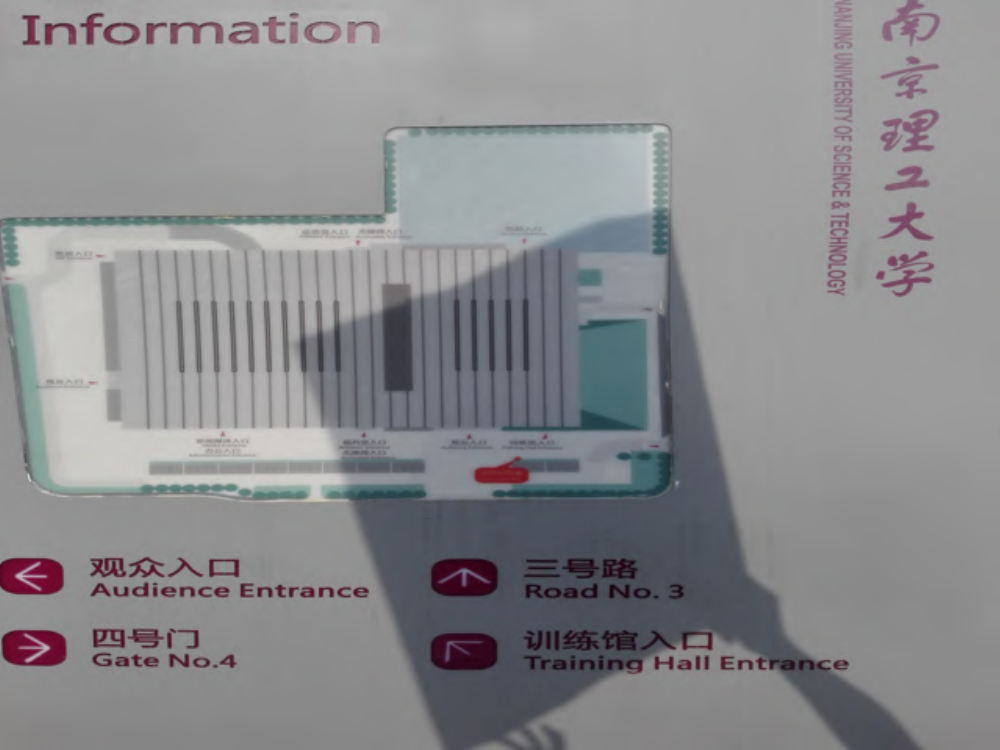}
        \end{minipage}
    }
    \hspace{-1em}
    \subfigure[\tiny{SADC}]{
        \begin{minipage}[ht]{0.22\linewidth}
            \includegraphics[width=\linewidth, height=\linewidth]{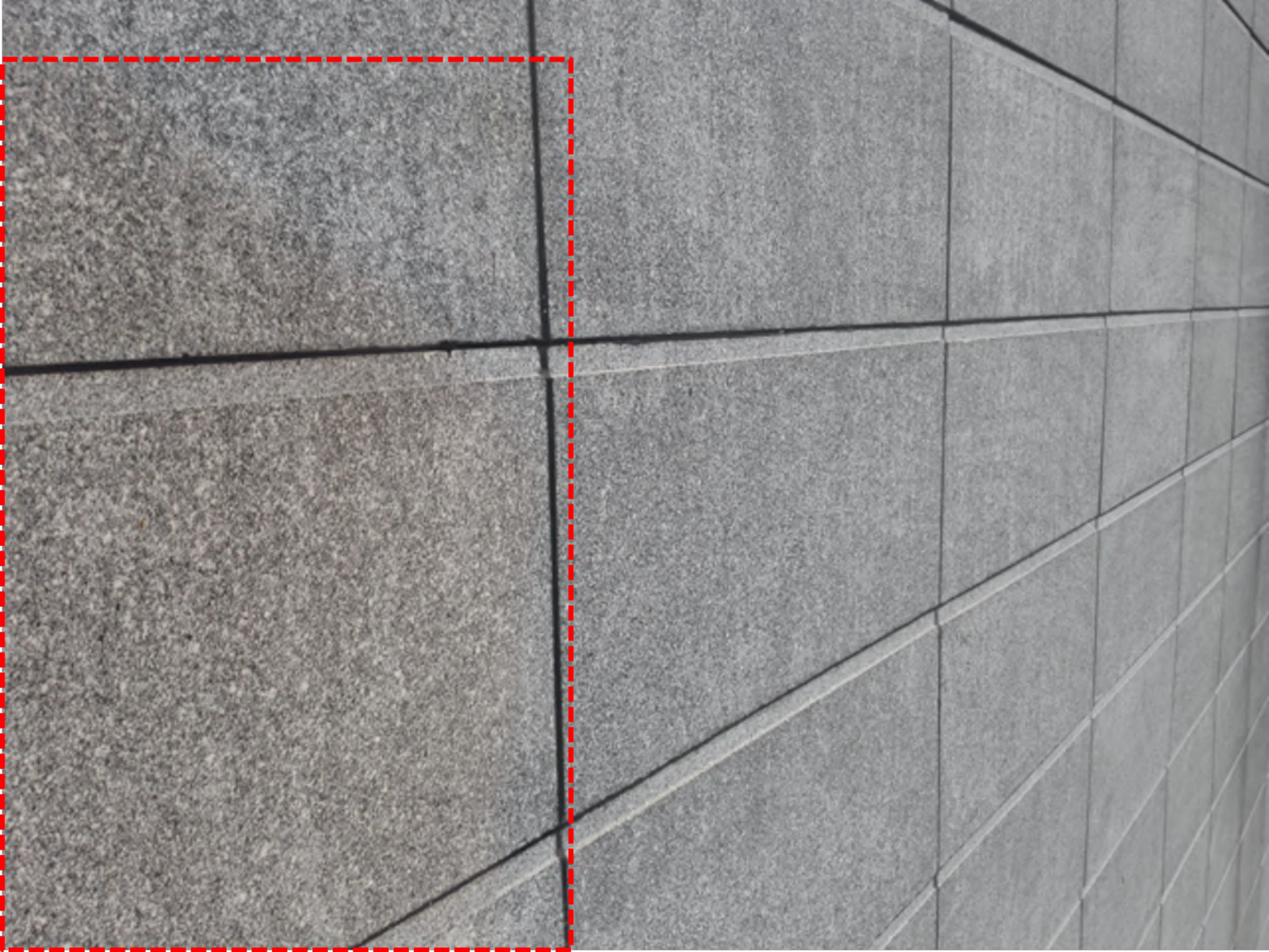} \\
            \includegraphics[width=\linewidth, height=\linewidth]{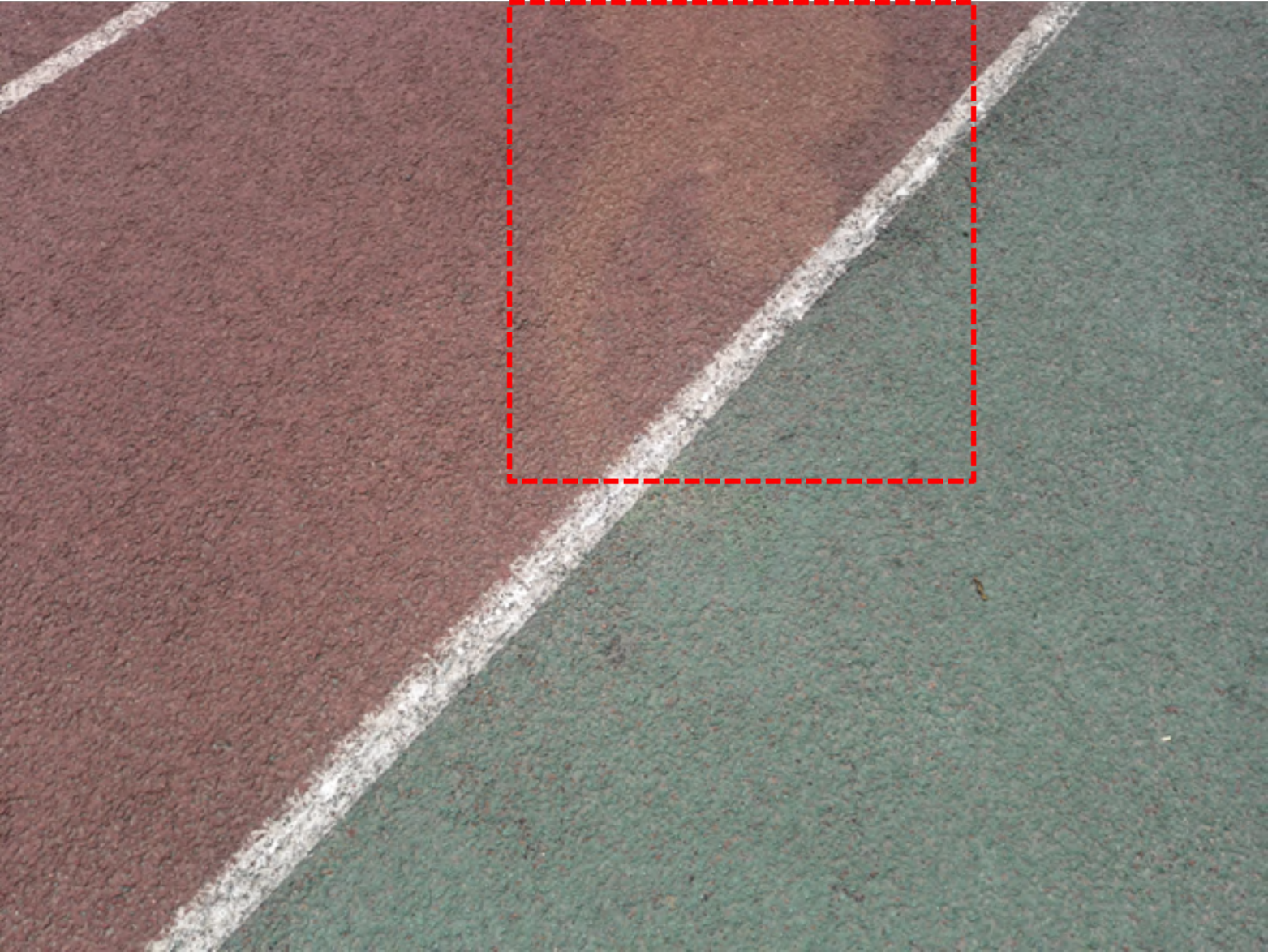} \\
            \includegraphics[width=\linewidth, height=\linewidth]{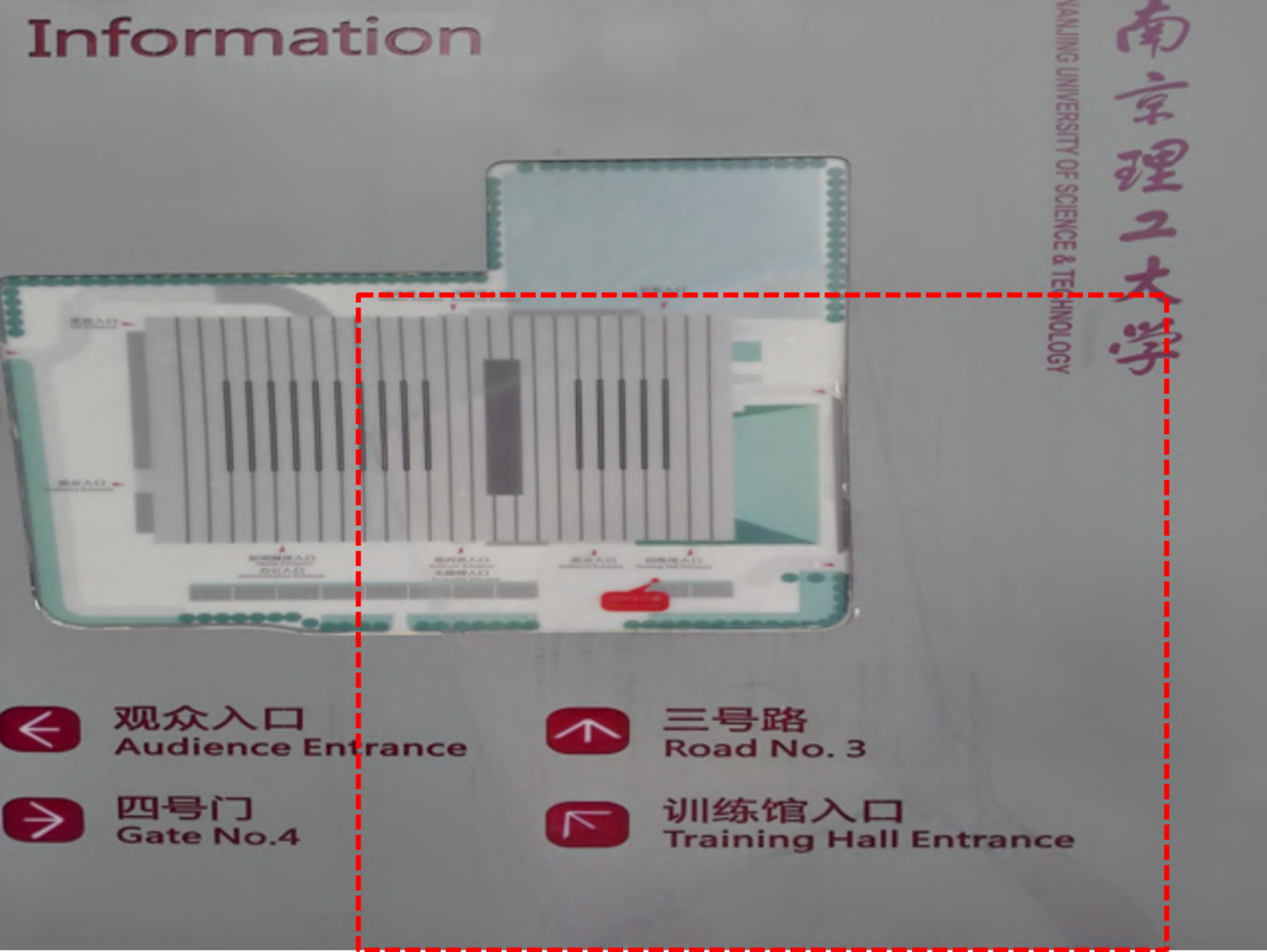}
        \end{minipage}
    }
    \hspace{-1em}
    \subfigure[\tiny{SADC+Distillation}]{
        \begin{minipage}[ht]{0.22\linewidth}
            \includegraphics[width=\linewidth, height=\linewidth]{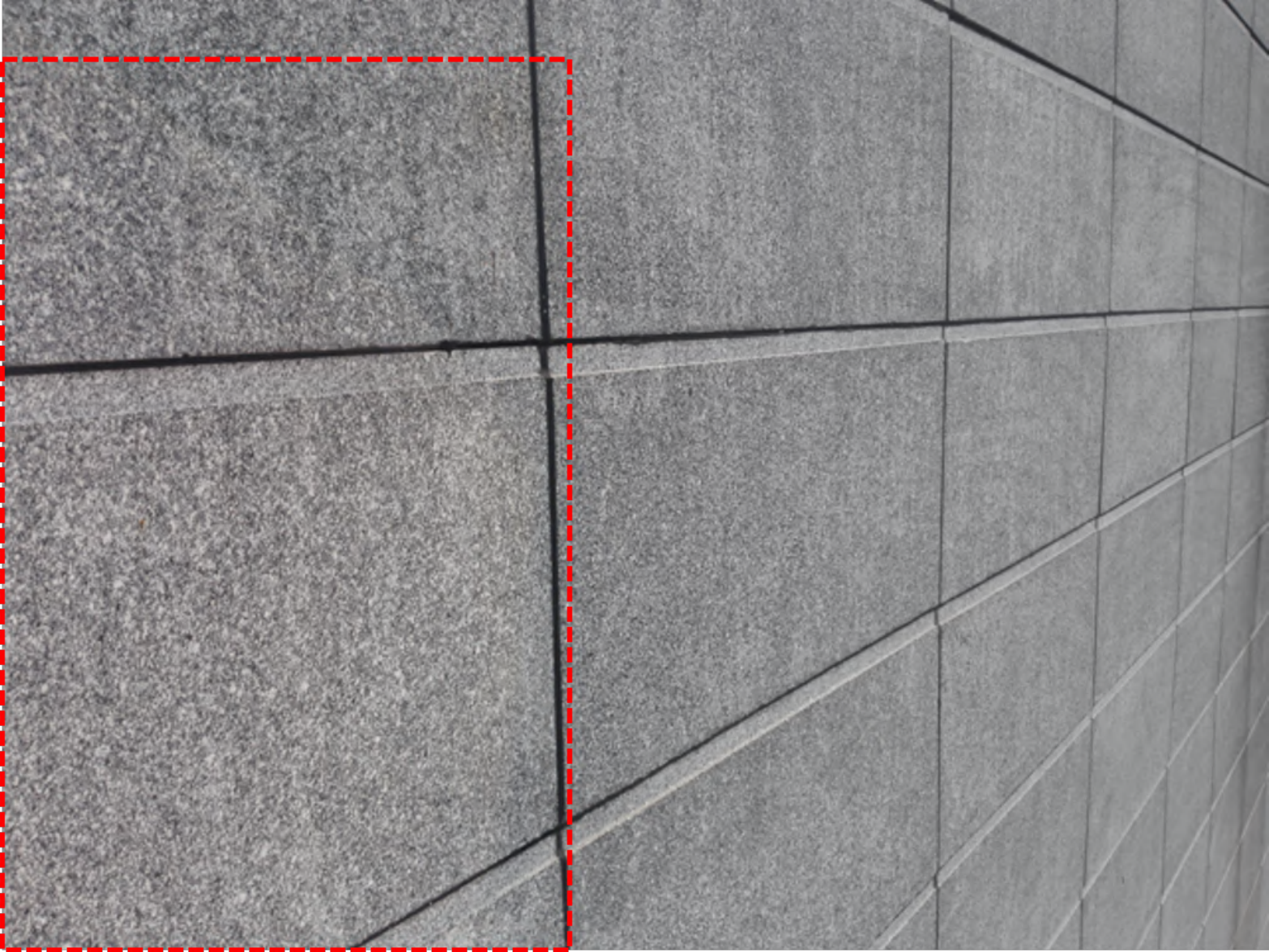} \\
            \includegraphics[width=\linewidth, height=\linewidth]{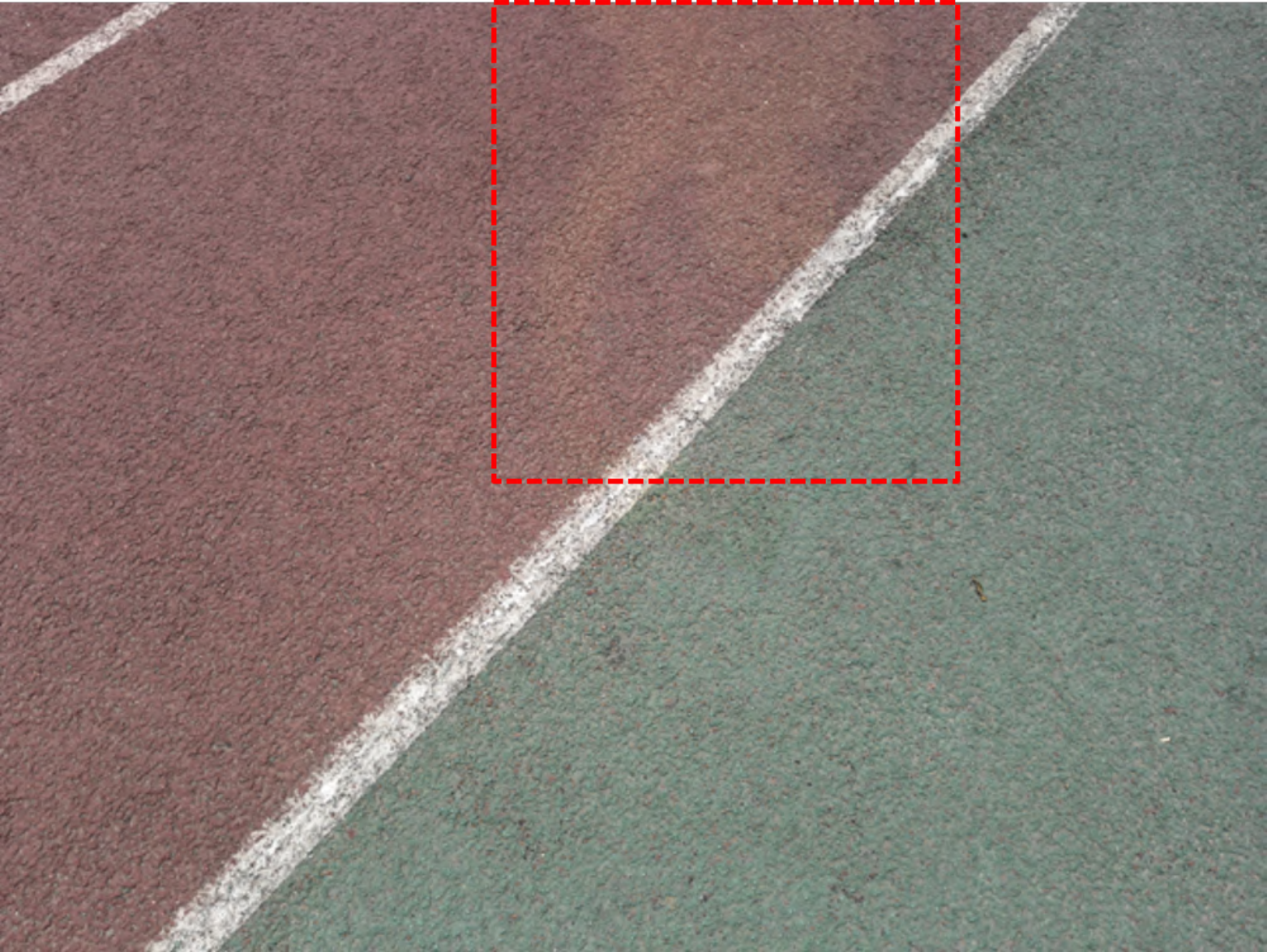} \\
            \includegraphics[width=\linewidth, height=\linewidth]{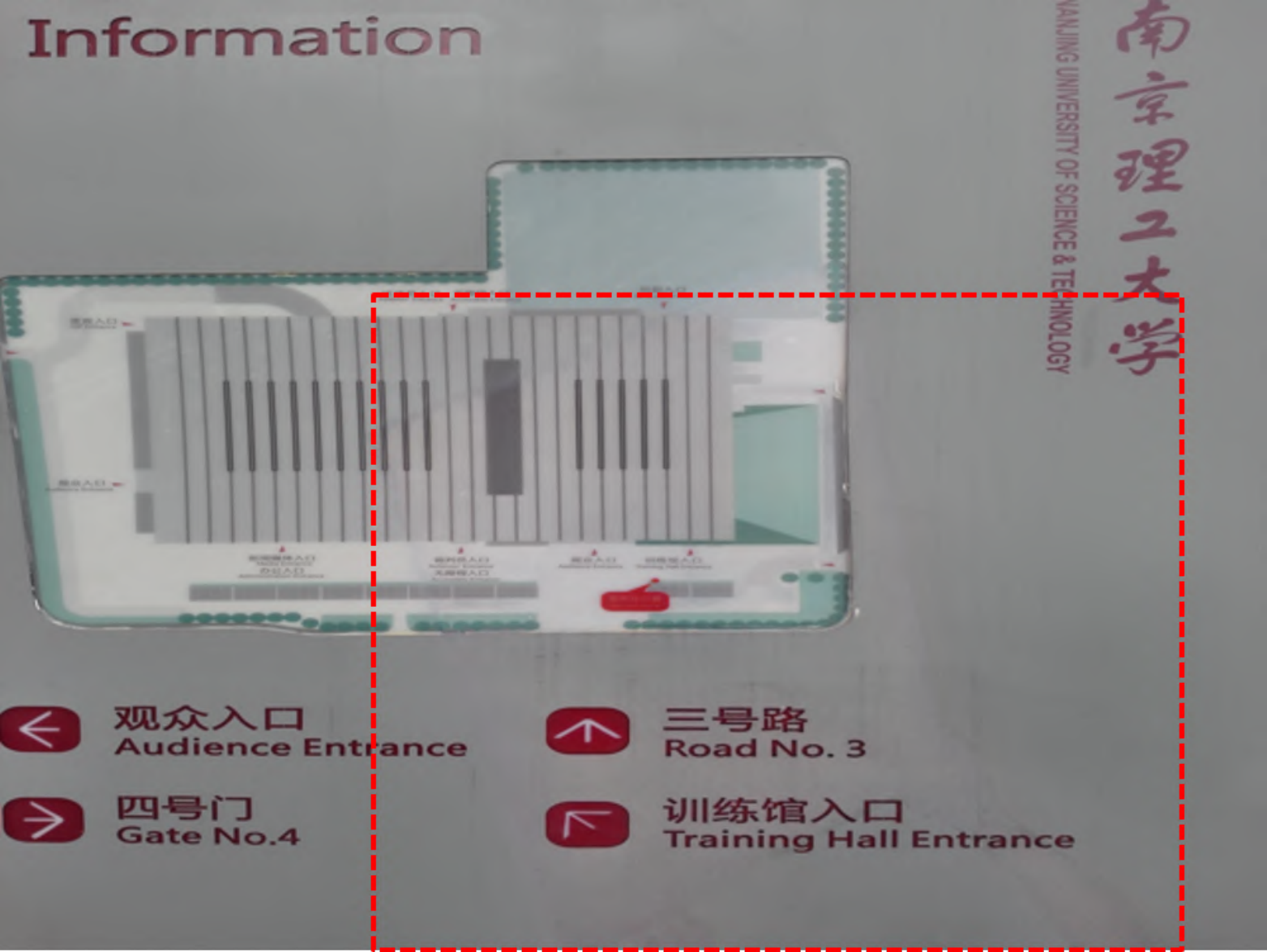}
        \end{minipage}
    }
    \hspace{-1em}
    \subfigure[\tiny{GT}]{
        \begin{minipage}[ht]{0.22\linewidth}
            \includegraphics[width=\linewidth, height=\linewidth]{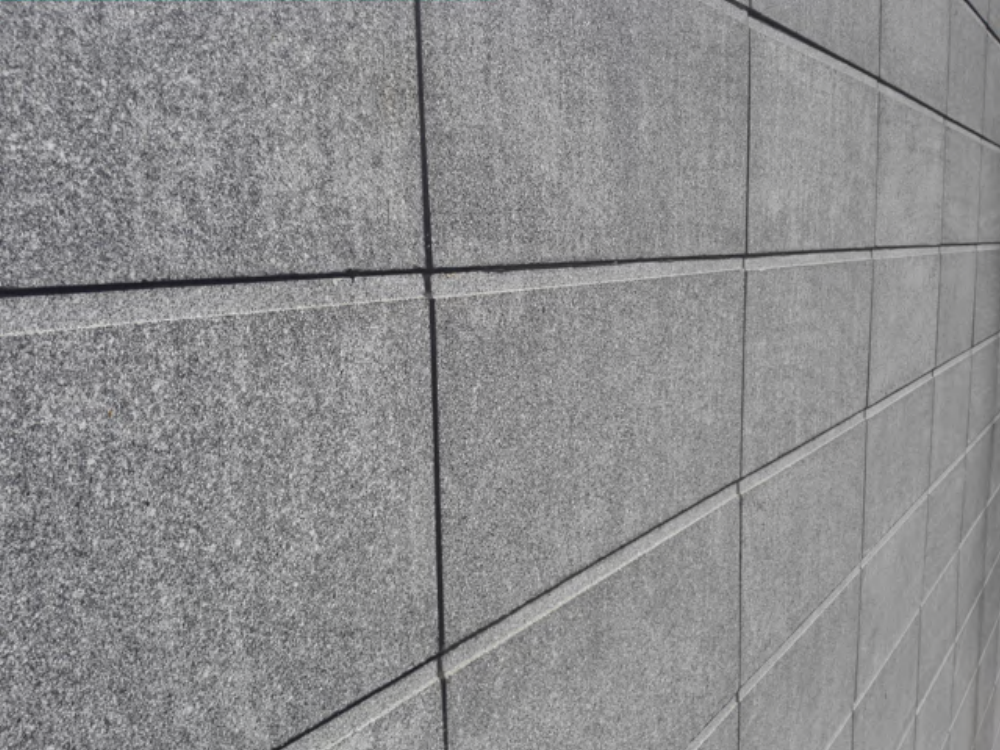} \\
            \includegraphics[width=\linewidth, height=\linewidth]{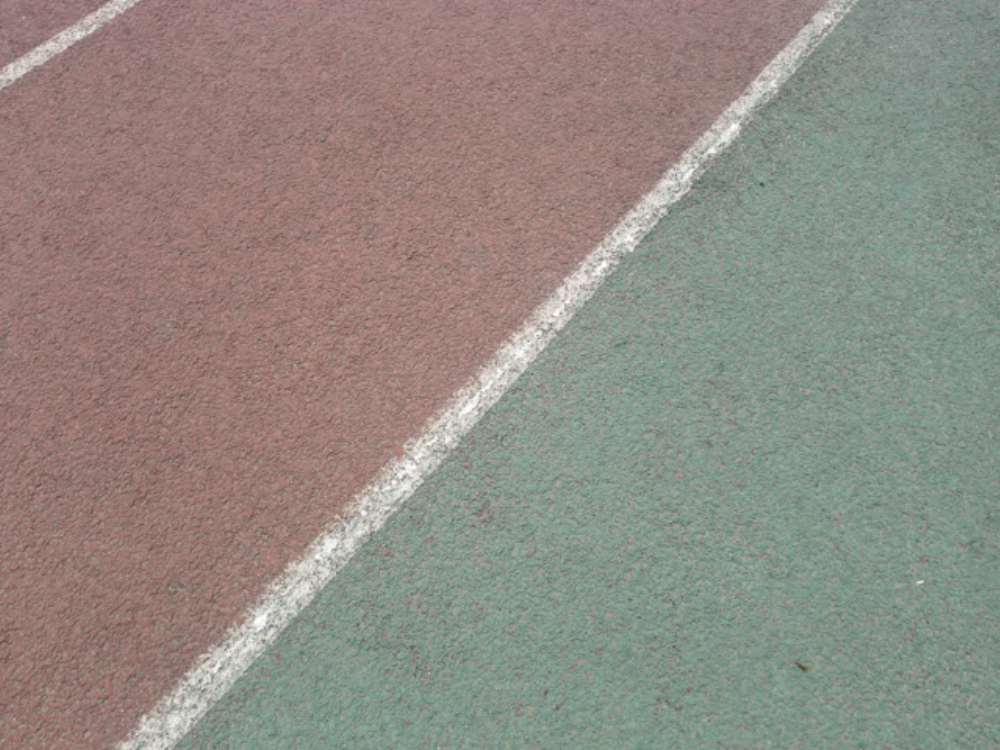} \\
            \includegraphics[width=\linewidth, height=\linewidth]{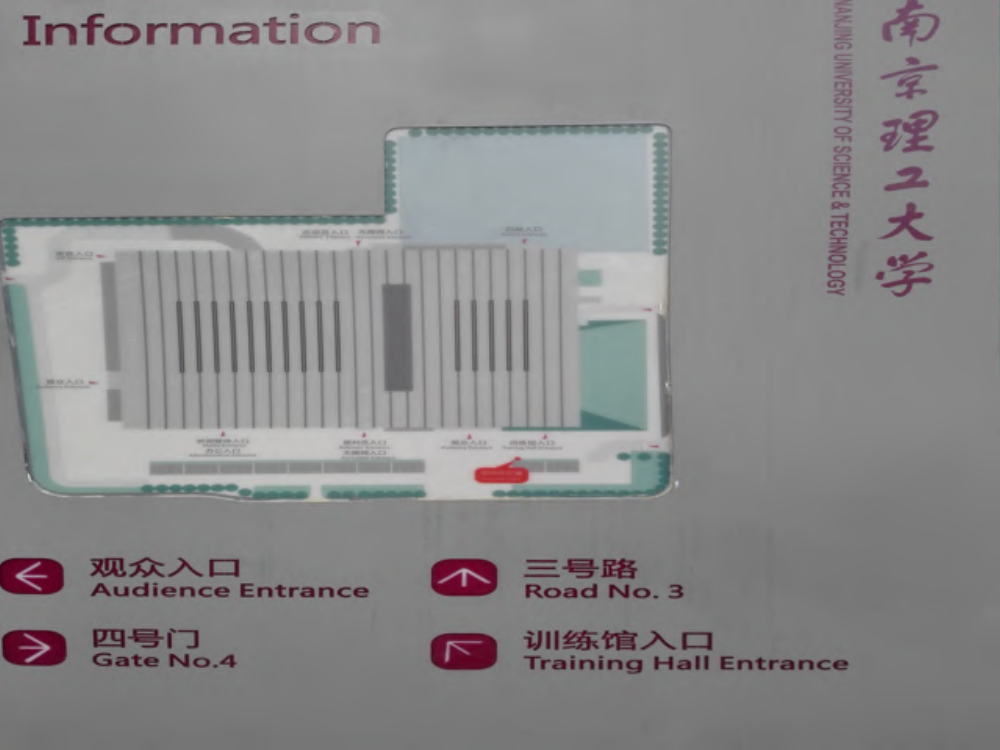}
        \end{minipage}
    }
    \caption{Qualitative comparison of shadow removal results among different combinations of our methods.}
    \label{figure:effective}
\end{figure}


\textbf{Component Efficacy}. In Table\,\ref{table:effective}, the effectiveness of different components in our method can be validated by either training solely with the normal convolution (\emph{i.e.}, baseline) or joint with the SADC module and intra-class distillation loss. By adopting the proposed SADC module, the RMSE of the shadow region decreases from 7.34 to 6.17, and the RMSE in the non-shadow region also decreases from 5.65 to 4.52. The results validate our assumption that the color mapping of the non-shadow region contradicts to that of the shadow region. Therefore, the separation of the non-shadow convolutions from the shadow convolutions is of vital importance for better shadow removal results. Furthermore, by employing the intra-class distillation loss, the RMSE in the shadow region shows a decrease of 0.17 by absorbing the nearby color result in the non-shadow region. The visualization comparison between the results either obtained by only adopting the SADC module or obtained by both the SADC module and the distillation loss is given in Fig.\,\ref{figure:effective}. The comparison between images in the second column and third column in the corresponding shadow region shows that the intra-convolution distillation loss can enhance the shadow removal results with extra supervision from the non-shadow region.

\begin{figure}[!t]
    \centering
    \includegraphics[width=\linewidth]{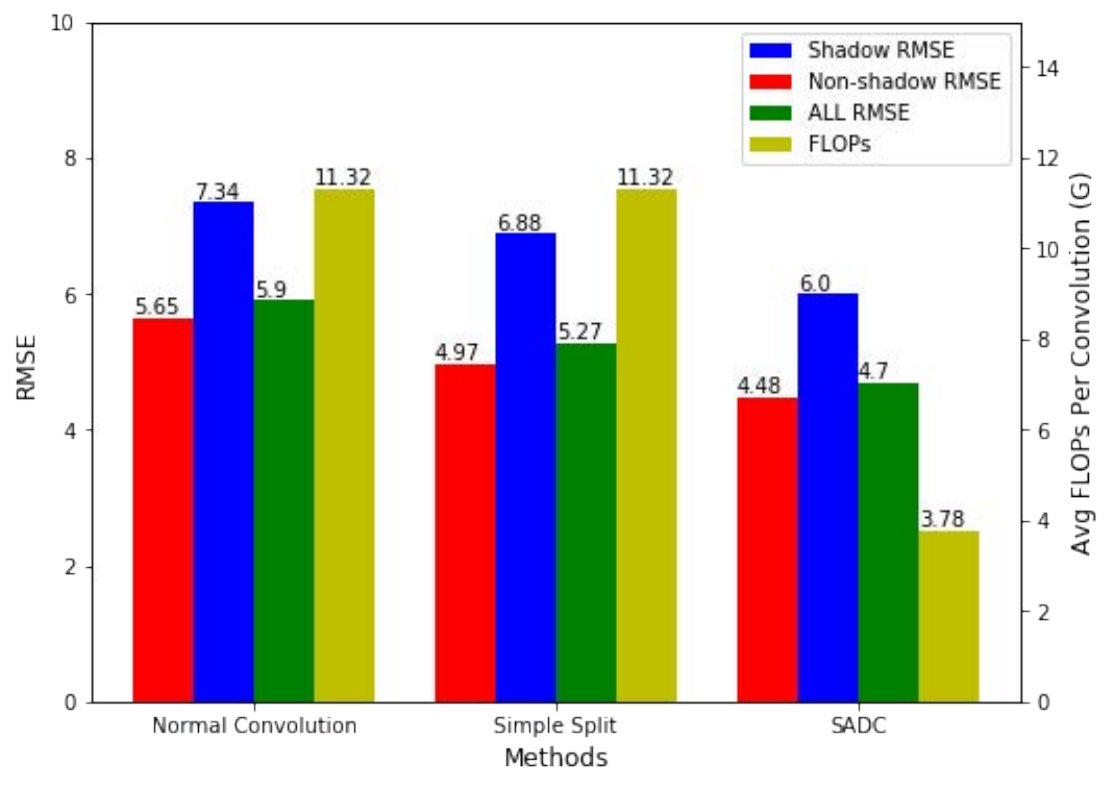}
    \caption{The comparison between different methods to validate the efficiency of the proposed SADC module.}
    \label{fig:efficiency}
\end{figure}

\textbf{SADC Efficiency}. As shown in Fig.\,\ref{fig:efficiency}, we fairly compare the proposed SADC module with a) the normal convolution and b) simply splitting the convolution into two branches (\emph{i.e.}, one for the shadow region and the other for the non-shadow region). Obviously, the simple separation of the convolution can decouple the interdependence between the shadow region and the non-shadow region. The RMSE in the non-shadow region decreases from 5.65 to 4.97. However, the RMSE in the shadow region does not show an abrupt reduction compared with that in the normal convolution. In our opinion, the color mapping in the shadow region is too sophisticated for a normal convolution to learn. This problem can be solved by the proposed SADC module with re-assignment of the computation resource corresponding to the shadow and non-shadow region. By adopting a lightweight convolution in the non-shadow branch, more complex convolution modules can be used to process the shadow region. 

The SADC module has better performance in both the shadow region (RMSE from 6.88 to 6.0) and the non-shadow region (RMSE from 4.97 to 4.48). Furthermore, the average FLOPs per convolution module also decreases from 11.32G to 3.78G in the SADC. Furthermore, the proposed SADC module can realize real acceleration compared with the normal convolution. The average inference time for a normal $3\times3$ convolution can reduce from 41.32ms to 13.68ms, which reaches around  3$\times$  acceleration. 

\begin{figure}[!t]
    \centering
    \includegraphics[width=\linewidth]{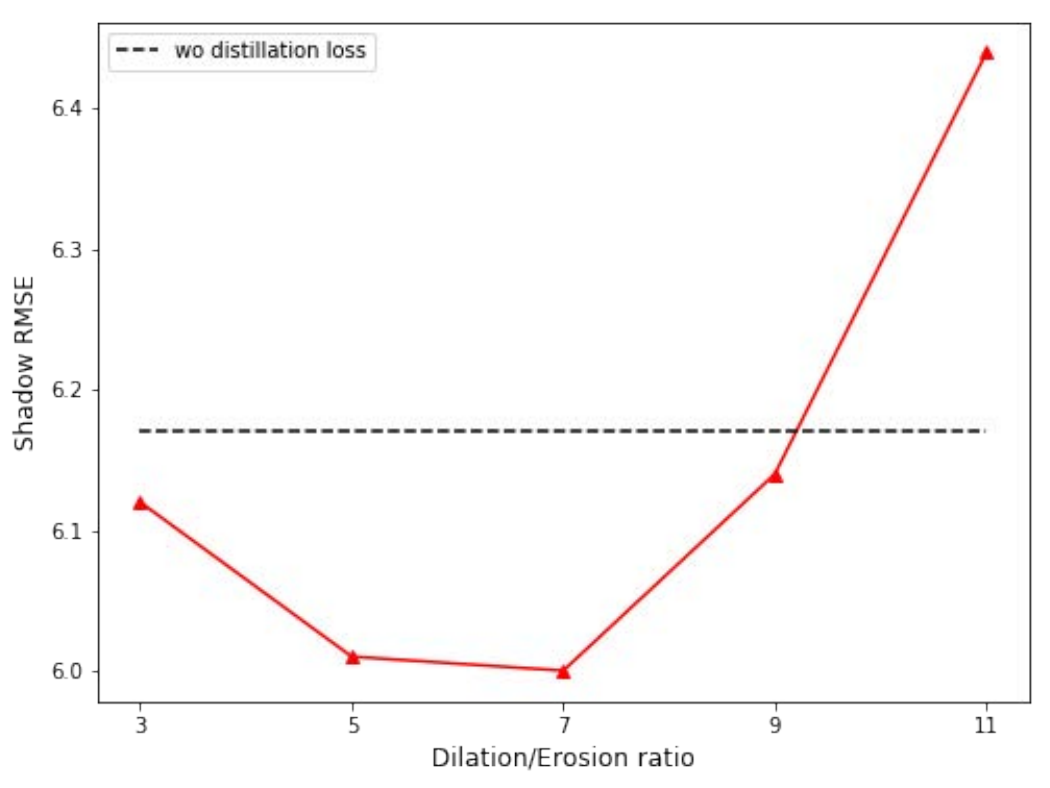}
    \caption{The influence of different dilation/Erosion ratio $\kappa$.}
    \label{fig:kappa}
\end{figure}

\textbf{Erosion and Dilation Ratio $\kappa$}.
The dilation ratio is an important hyper-parameter in the intra-convolution distillation loss. In Fig.\,\ref{fig:kappa}, a large dilation ratio $\kappa$ may contain more irrelevant color information, while a small dilation ratio is not sufficient to build a correlation between the shadow region and the non-shadow region. Hence, we adopt an intermediate dilation ratio $\kappa = 7$, which receives the best performance.

\begin{table}[htp]
\caption{Influence of different training strategy of $\lambda_2$.}
\resizebox{\linewidth}{!}{
\begin{tabular}{c|ccc}
\hline
Methods & \multicolumn{1}{c|}{Shadow RMSE} & \multicolumn{1}{c|}{Non-shadow RMSE} & ALL RMSE \\ \hline
Constant & 6.46 & 4.72 & 5.00 \\
Warmup and Constant & 6.43 & 4.70 & 4.96 \\
Warmup and monotonic increase & 6.00 & 4.48 & 4.70 \\ \hline
\end{tabular}}
\label{table:training_stg}
\end{table}

\textbf{Training Strategy}. To make better use of the intra-convolution distillation loss, we introduce a coefficient $\lambda_2$ which controls the influence level of the non-shadow region to the shadow region. Different training strategies can lead to different shadow removal results of the proposed intra-convolution distillation loss. Three strategies are proposed as, (1) Constant: Training with a constant coefficient (1 in our experiments); (2) Warmup and Constant: In the early epochs $\lambda_2 = 0$ and then $\lambda_2 = 1$; and (3) Warmup and monotonic increase: As shown in the Sec.\,\ref{overall}.

The results are shown in Table.\,\ref{table:training_stg}. As used in our method, the third strategy turns out to have better performance than the other two methods. In the early training stage, the result of the non-shadow region may produce misleading information and impede the overall shadow recovery results. By setting $\lambda_2 = 0$ in the early training stage, the shadow removal results become better as avoiding negative influence from inaccurate background information. With the training proceeds, the results of the non-shadow region can provide finer-grained background information to the shadow region to further enhance the shadow removal performance.

\textbf{Influence of Different Training Strategy.} As mentioned in Sec.\,\ref{shadow-aware}, training with the whole input feature maps for both the shadow branch and the non-shadow branch can a reach better convergence in the training phase. 

\begin{table}[htp]
\caption{Influence of different training strategy.}
\label{table:training_stg1}
\resizebox{\linewidth}{!}{
\begin{tabular}{c|ccc}
\hline
Methods & \multicolumn{1}{c|}{Shadow RMSE} & \multicolumn{1}{c|}{Non-shadow RMSE} & ALL RMSE \\ \hline
\textbf{Fashion 1} & 6.00 & 4.48 & 4.70 \\ 
\textbf{Fashion 2} & 6.15 & 5.18 & 5.30 \\ \hline
\end{tabular}}
\end{table}

To validate this, we conduct experiments by training the network with two fashions:
In \textbf{Fashion 1}, both shadow branch and non-shadow branch are fed with the whole input feature maps.
In \textbf{Fashion 2}, the input feature maps are separated into shadow region and non-shadow region first, and respectively fed to the shadow branch and non-shadow branch.
The results are shown in Table\,\ref{table:training_stg1}. As can be seen, \textbf{Fashion 1} performs consistently better than \textbf{Fashion 2}, well demonstrating the correctness of our training manner in the main paper.


\textbf{Generalization capability of SADC.} To validate the generalization capacity of the proposed SADC module, we use the pretrained model on the ISTD dataset and perform shadow removal on the SRD dataset.

\begin{table}[h]
    \caption{Generalization capability of SADC.}
    \centering
    \begin{tabular}{c|ccc}
    \hline
    Methods & \multicolumn{1}{c|}{Shadow RMSE} & \multicolumn{1}{c|}{Non-shadow RMSE} & ALL RMSE \\ \hline
    DHAN & 20.96 & 10.27 & 13.68\\
    DHAN\dag & 20.45 & 13.59 & 15.64 \\
    Ours & 19.84 & 8.47 & 11.64 \\ \hline
    \end{tabular}
    \label{Table:Generalize}
\end{table}

The results are shown in Tab.\,\ref{Table:Generalize}. Compared with DHAN and DHAN\dag, our method performs better when  trained on the ISTD dataset and tested on the SRD dataset, which demonstrates a better generalization capacity of the proposed method.

\section{Conclusion}
In this paper, we proposed a plug-and-play Shadow-Aware Dynamic Convolution (SADC) for shadow removal.
The SADC module separated the convolution operation between the shadow region and non-shadow region and re-assigned the computation cost based on the difficulty of learning a proper color mapping function.
For the non-shadow region, we adopted a lightweight convolution module to save computational cost, while more sophisticated convolution modules were adopted in the shadow region to guarantee the reconstruction results.
Furthermore, we proposed a novel intra-convolution distillation loss to encourage the output in the shadow branch to share a similar color distribution with that in the non-shadow branch.
Comprehensive experiments on various benchmark datasets demonstrated that the proposed method not only recovers the shadow images with more details but also maps the non-shadow counterpart better to its ground truth. 
As a plug-and-play module, the SADC can be easily combined with many widely-used CNNs to boost their performance in shadow removal.
In future, we plan to conduct SADC to solve more challenging shadow removal tasks, such as video shadow removal.

\ifCLASSOPTIONcaptionsoff
  \newpage
\fi



\bibliographystyle{IEEEtran}
\bibliography{main}
%

%
\end{document}